\documentclass[conference]{IEEEtran}
\IEEEoverridecommandlockouts
\usepackage{cite}
\usepackage{amsmath,amssymb,amsfonts}
\usepackage{algorithmic}
\usepackage{graphicx}
\usepackage{textcomp}
\usepackage{xcolor}
\usepackage{enumitem}
\usepackage{url}
\usepackage{tabularx}
\usepackage{ragged2e}
\usepackage{subcaption}
\usepackage{soul}
\usepackage{multirow}
\usepackage{booktabs}
\def\BibTeX{{\rm B\kern-.05em{\sc i\kern-.025em b}\kern-.08em
    T\kern-.1667em\lower.7ex\hbox{E}\kern-.125emX}}
\begin{document}

\title{Historically Relevant Event Structuring for Temporal Knowledge Graph Reasoning}

\author{
	\IEEEauthorblockN{
		Jinchuan Zhang\IEEEauthorrefmark{2}\textsuperscript{1}, 
		Ming Sun\IEEEauthorrefmark{2}\textsuperscript{2}*,
            Chong Mu\IEEEauthorrefmark{2}\textsuperscript{1}, 
            Jinhao Zhang\IEEEauthorrefmark{2}\textsuperscript{1},
		Quanjiang Guo\IEEEauthorrefmark{2}\textsuperscript{3},
		and Ling Tian\IEEEauthorrefmark{2}\textsuperscript{2}}
        \IEEEauthorblockA{\IEEEauthorrefmark{2}\textit{School of Computer Science and Engineering}, \textit{University of Electronic Science and Technology of China}, \textit{Chengdu}, \textit{China}\\
		\textsuperscript{1}\{jinchuanz, muchong, 202011090907\}@std.uestc.edu.cn, \textsuperscript{2}\{sunm, lingtian\}@uestc.edu.cn, \textsuperscript{3}guochance1999@163.com}
        \thanks{* Corresponding author: Ming Sun.}
}

\maketitle

\begin{abstract}
Temporal Knowledge Graph (TKG) reasoning focuses on predicting events through historical information within snapshots distributed on a timeline. Existing studies mainly concentrate on two perspectives of leveraging the history of TKGs, including capturing evolution of each recent snapshot or correlations among global historical facts. Despite the achieved significant accomplishments, these models still fall short of I) investigating the impact of multi-granular interactions across recent snapshots, and II) harnessing the expressive semantics of significant links accorded with queries throughout the entire history, particularly events exerting a profound impact on the future. These inadequacies restrict representation ability to reflect historical dependencies and future trends thoroughly. To overcome these drawbacks, we propose an innovative TKG reasoning approach towards \textbf{His}torically \textbf{R}elevant \textbf{E}vents \textbf{S}tructuring (HisRES). Concretely, HisRES comprises two distinctive modules excelling in structuring historically relevant events within TKGs, including a multi-granularity evolutionary encoder that captures structural and temporal dependencies of the most recent snapshots, and a global relevance encoder that concentrates on crucial correlations among events relevant to queries from the entire history. Furthermore, HisRES incorporates a self-gating mechanism for adaptively merging multi-granularity recent and historically relevant structuring representations. Extensive experiments on four event-based benchmarks demonstrate the state-of-the-art performance of HisRES and indicate the superiority and effectiveness of structuring historical relevance for TKG reasoning.
\end{abstract}


\section{Introduction}
Temporal Knowledge Graphs (TKGs) \cite{tkgsurvey} have revolutionized the limitations of static KGs by considering temporal information. Specifically, TKGs advance beyond traditional triple facts, enhancing them into (\textit{subject}, \textit{relation}, \textit{object}, \textit{timestamp}) quadruples, thereby bridging the divide between static and dynamic characteristics. This enrichment facilitates capturing the dynamic evolution of real-world events and holds significant practicality. It expands the downstream applications to include event prediction \cite{zhouwt}, temporal question answering \cite{llmtkgr}, social network analysis \cite{snet}, anomaly detection \cite{Iot}, \textit{etc}.

As established in prior research \cite{regcn, tirgn}, TKGs manifest in the form of sequential snapshots or subgraphs partitioned along a timeline, where each snapshot encapsulates concurrent events occurring at specific timestamps, thereby providing an intuitive representation of event evolutionary patterns. However, TKGs persistently suffer from incompleteness and encounter unique challenges attributable to their temporal characteristics. The critical task of addressing these limitations can be categorized into interpolation and extrapolation, involving reasoning missing facts (events) at past and future timestamps according to historical knowledge, respectively.

From the perspective of history utilization, extrapolation methods can be classified into two principal categories: a) tallying statistics of historical events and leveraging repetitive patterns \cite{cygnet,cenet}; b) capturing concurrent and evolutionary correlations from the most recent snapshots \cite{renet,regcn,cen,RETIA,rpc,l2tkg}. In detail, the a) category generates sparse matrices from repetitive statistics to mask predictions. The b) category adopts Graph Neural Networks (GNNs) to model structural dependencies among concurrent facts, and Recurrent Neural Networks (RNNs) to capture the sequential evolution of snapshots. 
Based on the b) category, few innovative models \cite{tirgn,hgls,logCL} are proposed to leverage global historical information. Among them, TiRGN \cite{tirgn} employs a global historical encoder to constrain the prediction scope. HGLS \cite{hgls} establishes connections between identical entities across different timestamps to capture multi-hop correlations among globally historical events. LogCL \cite{logCL} constructs a graph consisting of historical facts relevant to queries to obtain rich semantics.

Despite these advancements, previous studies still fall short of comprehensively exploring historically relevant events in TKGs. This limitation stems from two specific challenges: I) \textit{\textbf{Recent Event Impact}}: investigating the impact of multi-granular interactions across recent snapshots, and II) \textit{\textbf{Distant Historical Influence}}: harnessing expressive semantics of significant links accorded with queries throughout the entire history, particularly events exerting profound future impact.

I) \textit{\textbf{Recent Event Impact}}: the first noteworthy challenge arises from inadequate explicit and flexible assessment of multi-granular correlations between facts across recent adjacent snapshots. Recent facts typically exert stronger influence on future events and can initiate chains of events spanning multiple temporal intervals. As sequentially occurring events at adjacent timestamps often share causal relations, the aforementioned deficiency results in the loss of crucial inter-event dependencies. Although RPC \cite{rpc} and LogCL \cite{logCL} incorporate inter-snapshot information and assign variable weights to differentiate contributions of recent snapshots, they remain limited to the individual snapshot level, potentially overlooking fact-level propagation patterns that provide critical signals for effective TKG reasoning.

II) \textit{\textbf{Distant Historical Influence}}: the second significant challenge concerns temporally distant historical facts that can substantially influence future events despite their remoteness. Existing works frequently overlook crucial information intrinsic in globally relevant event structures, particularly long-term semantic patterns and periodic interactions, resulting in inadequate representations. While HGLS \cite{hgls} attempts to address this by connecting identical entities across snapshots through additional edges, it incorporates redundant information from distant timestamps and requires excessive computational complexity. LogCL \cite{logCL} integrates query-relevant historical facts but fails to explicitly prioritize key historical links during aggregation, thereby limiting its ability to differentiate the relative importance of historical facts from various timestamps.
\vspace{-12pt}
\begin{figure}[h]
\centerline{\includegraphics[width=1\linewidth]{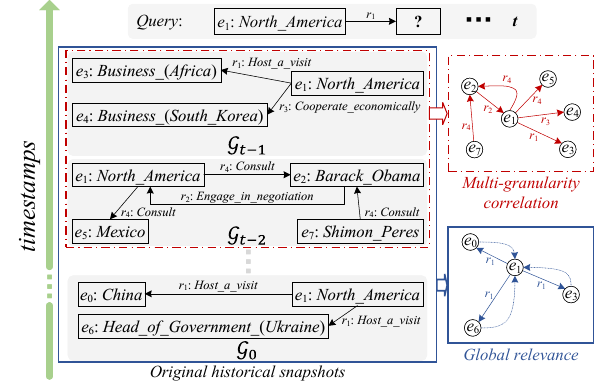}}
\caption{Example of extrapolation over TKGs. Left: original historical snapshots containing timestamped facts. Right: derived graphs structured through multi-granularity correlation (red elements) and global relevance (blue elements) perspectives.}
\label{example}
\end{figure}

To tackle these aforementioned challenges, this paper classifies them into two specific interactions within TKGs: multi-granularity correlation and global relevance. Figure \ref{example} exemplifies these concepts using an instance from ICEWS \cite{icews}. Each entity is assigned a unique numeric identifier (e.g., $e_1$ denotes \textit{North}\_\textit{America}). For clarity, the entities on the right side of Figure \ref{example} are represented as simplified nodes.
\begin{itemize}[leftmargin=*]
    \item Multi-granularity correlation (denoted by red dotted frame) captures inter-snapshot correlations by structuring facts across varying span of recent timestamps, rather than complex evolutionary patterns. The nodes and edges (red arrow) represent the aggregation of all facts from two adjacent snapshots. Such structure enables the capture of lasting effects through two-hop paths across consecutive timestamps (e.g., $e_2\rightarrow r_2 \rightarrow e_1 \rightarrow r_1 \rightarrow e_3$, where $(e_2, r_2, e_1)$ occurs at $t-2$, and $(e_1, r_1, e_3)$ at $t-1$). This addresses the challenge of I) \textit{\textbf{Distant Historical Influence}} by highlighting the interplay between facts in the most recent snapshots.
    \item Global relevance (denoted by blue solid frame) encompasses significant facts related to the query ($e_1$, $r_1$) by structuring links (blue arrow) throughout the original historical snapshots. This addresses the challenge of II) \textit{\textbf{Distant Historical Influence}}, emphasizing key events that hold profound implications for future from distant timestamps among numerous historical events. From a global relevance perspective, the occurrence of $(e_1, r_1, e_0)$ at $t=0$ may influence $(e_1, r_1, e_3)$ at $t-1$ through two-hop path connections.
\end{itemize}

Focusing on these two interactions, this paper proposes a novel method that specializes in TKG reasoning with \textbf{His}torically \textbf{R}elevant \textbf{E}vent \textbf{S}tructuring (HisRES). Specifically, HisRES adopts an encoder-decoder architecture, focusing on two critical aspects of TKG reasoning: capturing the structural and temporal patterns within recent snapshots and understanding the impact of relevant events throughout the entire history. In terms of recent history, HisRES employs a multi-granularity evolutionary encoder (\S \ref{sec:mee}), which models concurrent interactions and evolution within each snapshot and further integrates adjacent snapshots to learn temporal dependencies using multi-hop links (addressing the first challenge). To explicitly capture structural correlations across the entire history (addressing the second challenge), HisRES introduces a global relevance encoder, structuring historical facts directly relevant to the query set. We propose an innovative convolution-based graph attention network (ConvGAT) to efficiently emphasize important links of historical relevance for future prediction. Furthermore, HisRES incorporates a self-gating mechanism (\S \ref{sec:sgm}) to adaptively fuse entity representations from different granularities or encoders, achieving a comprehensive understanding by considering various historical information (i.e., global and local, intra-snapshot and inter-snapshot) of TKGs, leading to accurate predictions. Extensive experiments on four benchmarks demonstrate the superiority and effectiveness of HisRES in TKG reasoning, particularly in structuring historical relevance.

In general, the main contributions of this paper can be summarized as follows:
\begin{itemize}[leftmargin=*]
    \item We propose a novel TKG reasoning method termed HisRES, which sufficiently discovers structural and temporal dependencies through historically relevant event structuring.
    \item Through the adaptive fusion of multi-granularity evolutionary and global relevance encoding, HisRES can effectively integrate representations throughout historical interactions.
    \item Extensive experiments on four event-based datasets demonstrate that HisRES outperforms state-of-the-art TKG reasoning baselines, showcasing its effectiveness and superiority.
\end{itemize}
\section{Related Work}
KG reasoning \cite{tian} has been extensively studied in recent years, making significant contributions to downstream applications. Traditional reasoning deals mainly with static KGs such as the widely used FB15k-237 \cite{fb15k} and WN18RR \cite{conve}. In contrast, TKG reasoning is tailored for time-sensitive data such as ICEWS \cite{icews}, adept at capturing evolutionary patterns, and discovering future facts within event-based datasets characterized by temporal features. Our related work encompasses both static and temporal knowledge graph reasoning.
\subsection{Static Knowledge Graph Reasoning}
Early research on static KGs focuses on modeling entities and relations in various representation spaces. TransE \cite{transe} introduces a simple translation assumption $s + r \approx o$, while DistMult \cite{distmult} learns relational semantics through embeddings from a bilinear objective. In the complex vector space, ComplEx \cite{complex} and RotatE \cite{rotate} employ Hermitian dot products and relation rotations, respectively. With the advent of convolutional neural networks, approaches such as ConvE \cite{conve} and ConvTransE \cite{convtranse} utilize convolutional operations to fuse entity-relation embeddings. Recent advancements aim at learning more expressive structures with Graph Convolutional Networks (GCN) \cite{gcn}. R-GCN \cite{rgcn} extends GCNs with relation-specific block-diagonal matrices. CompGCN \cite{compgcn} integrates multi-relational GCN methods with various composition operations. To differentiate node contributions, GAT \cite{gat} employs attention mechanisms, while KBGAT \cite{KBGAT} uses relation-specific attention to capture neighboring semantics. However, these static approaches exhibit critical limitations: they neither effectively incorporate explicit temporal information nor adequately explore complex periodicity and historical influences. These deficiencies significantly constrain their ability to handle dynamic facts and learn evolutionary patterns.

\subsection{Temporal Knowledge Graph Reasoning}
TKG reasoning can be broadly categorized into two types based on the occurrence moment of predicted events: interpolation and extrapolation. The former deals with known timestamps, while the latter addresses future timestamps. Additionally, a few methods \cite{tilp,teilp,tilr} aim to predict time through inductive reasoning based on logical rules. Notably, this paper focuses on the extrapolation.

Assuming we know all facts before $t_T$ (including $t_T$), the extrapolation task aims to predict future facts at timestamps $t>t_T$ based on historical TKG sequences. Various approaches have been explored in recent years to address it. xERTE \cite{xerte} designs a subgraph sampling module based on temporal information and incorporates time within relation embeddings. GHNN \cite{ghnn} proposes graph hawkes neural network to capture the evolution of graph sequences. TANGO \cite{tango} extends neural ordinary differential equations to model dynamic knowledge, enabling continuous-time representation. TITer \cite{titer} and CluSTER \cite{cluster} employ reinforcement learning to generate query-specific paths composed of historical entities. TLogic \cite{TLogic} incorporates symbolic representations and learns temporal logical rules to enhance reasoning. 

Besides, numerous studies can be categorized into two main groups: those that focus on repeating statistical historical facts, and those that model the interactions among historical events. In the first category, CyGNet \cite{cygnet} employs a multi-hot vector as a historical vocabulary to record past facts, subsequently using this information to mask predictions. Building upon this, CENET \cite{cenet} and PLEASING \cite{pleasing} introduce contrastive learning and learnable classifiers to differentiate historical and potential events. In the latter category, RE-NET \cite{renet} and RE-GCN \cite{regcn} employ GCNs to model concurrent structures and utilize recurrent units to learn entity or relation evolution patterns. On this basis, CEN \cite{cen} considers historical length-diversity and time-variability. TiRGN \cite{tirgn} adopts a global history encoder like CyGNet to identify entities related to historical or unseen facts and a time-guided decoder. RETIA \cite{RETIA} and RPC \cite{rpc} introduce the concept of line graphs to learn the interaction of relations. LMS \cite{lms} learns entity and time interactions separately from a multi-graph perspective. L$^2$TKG \cite{l2tkg} addresses the missing associations and creates links between unconnected entities to obtain more comprehensive representations. Some novel insights attempt to capture global structural correlations through additional graphs. HGLS \cite{hgls} explicitly captures long-term dependencies by connecting the same entity at each timestamp, while LogCL \cite{logCL} proposes an entity-aware attention mechanism to flexibly adopt facts relevant to queries and utilizes contrastive learning to improve robustness.

Nevertheless, existing methods still face limitations in capturing two key interactions: I) \textit{\textbf{Distant Historical Influence}} and II) \textit{\textbf{Distant Historical Influence}}. Regarding the former, although RPC and LogCL propose mechanisms to calculate the correspondence of recent timestamps and distinguish the importance of different recent snapshots respectively, they fail to explicitly model the impact propagation of individual facts from the most recent snapshots—a capability crucial for predicting future repetitive or unseen facts. Concerning the latter, TiRGN employs a fixed history mask matrix to focus on historical repetitive facts, but this constrains the model's predictive capability as it fails to learn information from distant historical data. While HGLS establishes edges between identical entities at different timestamps to capture historical associations through multi-hop relations, it significantly multiplies the graph's node count and necessitates a multi-layer to capture multi-hop information, leading to increased computational overhead. Additionally, LogCL suffers from inefficient encoding when aggregating global information, making it difficult to differentiate the contributions of historical facts from different temporal periods.
\section{Method}
\begin{figure}[htbp]
\centerline{\includegraphics[width=\linewidth]{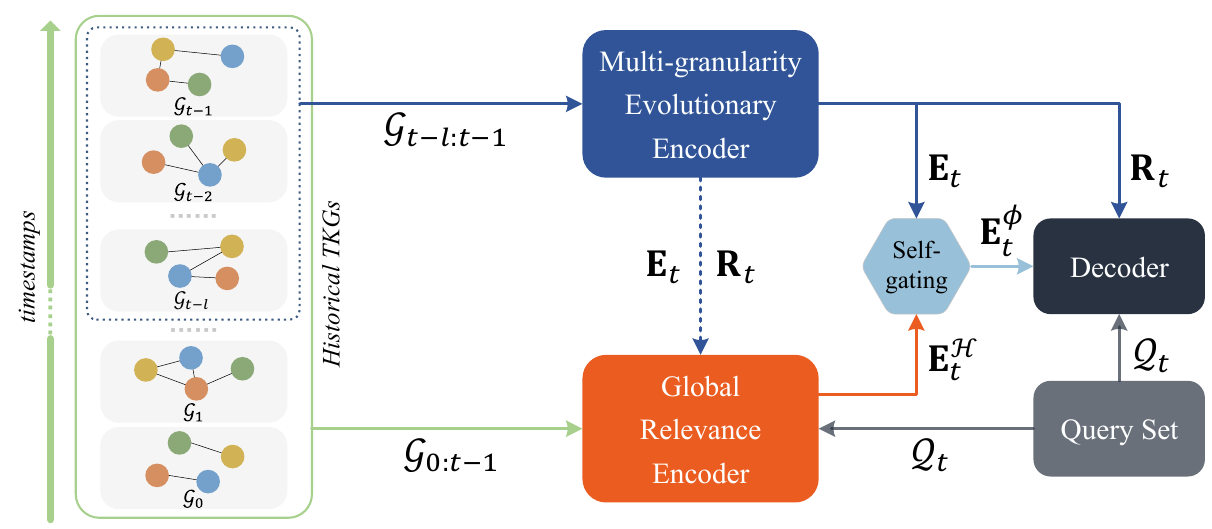}}
\caption{An illustration of HisRES model architecture.}
\label{framework}
\end{figure}

The overarching framework of HisRES is depicted in Figure \ref{framework}, which follows encoder-decoder architecture and consists of three key modules: i) \textit{Multi-granularity Evolutionary Encoder} (\S \ref{sec:mee}), which focuses on capturing correlations among concurrent facts in the most recent snapshots and modeling temporal evolution of these snapshots; ii) \textit{Self-gating Mechanism} (\S \ref{sec:sgm}), which studies a learnable weight for each entity and adaptively merges entity representations from different encoders; iii) \textit{Global Relevance Encoder} (\S \ref{sec:gre}), which integrates all historically relevant facts, and aggregates their crucial neighboring semantics with a novel graph attention network. In \textit{Decoder} (\S \ref{dec}), we employ a widely used ConvTransE.
\subsection{Notations}
In this paper, Temporal Knowledge Graphs (TKGs) are conceptualized as sequences of subgraphs, referred to as snapshots, and denoted as $G = \{\mathcal{G}_1, \mathcal{G}_2, ..., \mathcal{G}_{|\mathcal{T}|}\}$. Here, $G$ denotes the set of all TKG snapshots, $|\mathcal{T}|$ represents the total number of historical timestamps, and each snapshot $\mathcal{G}_t$ corresponds to the collection of facts or events occurring at timestamp $t$. Given $\mathcal{G}=\{\mathcal{E}, \mathcal{R}, \mathcal{F}, \mathcal{T}\}$, where $\mathcal{E}$, $\mathcal{R}$, and $\mathcal{F}$ represent the sets of entities, relations, and facts, respectively. Each fact $f \in \mathcal{F}$ is defined as a quadruple $(s, r, o, t)$, where $s, o\in\mathcal{E}$, representing the subject and object entities, $r \in \mathcal{R}$ indicates the relation, and $t \in \mathcal{T}$ is the timestamp of the quadruple. Given query set $\mathcal{Q}_t$ at timestamp $t$, all historical facts correlated with queries like $\{q=(s, r, ?, t)|q\in\mathcal{Q}_t\}$ before $t$, can be grouped within a global graph $\mathcal{G}^{\mathcal{H}}_t$ in HisRES when making predictions. Bold letters are used to indicate vectors. In summary, the important notations used in equations are presented in Table \ref{notations}.

\begin{table}[ht]
\caption{The summary of important notations.}
\resizebox{\linewidth}{!}{
\begin{tabular}{ll}
\toprule
Notations & Descriptions \\ 
\midrule
$\mathcal{E}$, $\mathcal{R}$, $\mathcal{T}, \mathcal{Q}$  & entity, relation, timestamp, and query set of TKGs\\
$\mathcal{G}_t$ & temporal knowledge graph snapshot at $t$ \\
$\textbf{E}_t$, $\textbf{R}_t$ & entity and relation embeddings of local evolution at $t$\\
$\mathcal{G}^{\mathcal{H}}_t$ & globally relevant graph containing all facts before $t$ \\
$\textbf{E}^{\mathcal{H}}_t$, $\textbf{R}^{\mathcal{H}}_t$ & entity and relation embeddings of global relevance at $t$\\
$\textbf{s}, \textbf{r}, \textbf{o}$   & entity and relation embeddings of a triple\\
$\Theta$ & self-gating vectors for embedding fusion\\
$\alpha$ & trade-off hyper-parameter for prediction task coefficient \\
$l$ & hyper-parameter to control local historical length\\
$\omega$ & hyper-parameter to control local granularity span\\
\bottomrule
\end{tabular}
}
\label{notations}
\end{table}

\textbf{Task Definition.} Our primary focus is on the extrapolation task within TKG reasoning, also known as link prediction problem at future timestamps.

\textbf{Formal Problem Formulation.} For a query $q \in \mathcal{Q}$ of the form $q = (s, r, ?, t)$ or $q = (?, r, o, t)$, the formal problem formulation of HisRES is to calculate the conditional probability of an object $o$ or subject $s$ given $(s, r, t)$ or $(r, o, t)$ and historical snapshots. These probabilities are denoted $p(o|s, r, t, \mathcal{G}_{0:t-1})$ and $p(s|r, o, t, \mathcal{G}_{0:t-1})$, respectively. It is noteworthy that we utilize object prediction as the illustrative example in our methodology section (\S \ref{lo}).
\subsection{Multi-granularity Evolutionary Encoder} \label{sec:mee}
\begin{figure}[htbp]
\centerline{\includegraphics[width=1\linewidth]{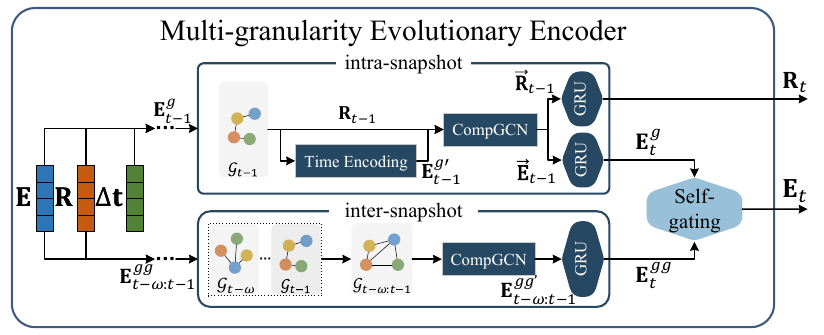}}
\caption{The achitecture of proposed multi-granularity evolutionary encoder.}
\label{mee}
\end{figure}

Multi-granularity evolutionary encoder serves as the cornerstone of HisRES, which entails predicting future events using historical snapshots from the most recent past. In accordance with this principle, we consider the most neighboring $l$ snapshots as historical data and aggregate features from both structural and temporal perspectives in the multi-granularity span. It includes intra-snapshot and inter-snapshot dependencies, the latter of which can structure several adjacent snapshots simultaneously (we select every $\omega$ adjacent snapshots). The illustration of one cycle in multi-granularity evolutionary encoder is depicted in Figure \ref{mee}.
\subsubsection{Structural and Temporal Modeling intra-snapshot}
The fundamental component of HisRES involves modeling structural and temporal features from each recent snapshot, encompassing all characteristics of TKGs (i.e., entity, relation and time information), which is shown in the \textit{intra-snapshot} of Figure \ref{mee}.

\paragraph{Time Encoding}
Analyzing the evolution process of events, periodicity and time attributes are crucial factors. Thus, similar to previous efforts \cite{hismatch,logCL}, we encode the time information $\Delta\textbf{t}$ based on the time interval between the current time $t_i$ and the predicted time $t$, as follow:
\begin{equation} \label{te1}
    \Delta\textbf{t}=\sigma(\textbf{w}_t(t-t_i)+\textbf{b}_t){,}
\end{equation}
where $\sigma$ is the cosine periodic activation function, $\textbf{w}_t$ and $\textbf{b}_t$ are the learnable parameter and bias of time. Hence, the initial entity embeddings for each single snapshot can be calculated as:
\begin{equation} \label{te2}
    \textbf{E}^{g'}_{t-1} =\textbf{W}_0([\textbf{E}^{g}_{t-1}||\Delta\textbf{t}]){,}
\end{equation}
where $\textbf{W}_0\in\mathbb{R}^{d \times 2d}$ is a linear transformation. In particular, all our modules use $t-1$ as a reference time.

\paragraph{Entity Aggregation}
In the structural aspect, the basic aggregation resembles that of CompGCN \cite{compgcn}. To merge entities and relations effectively, we employ composition operators such as "\textit{subject}+\textit{relation}", akin to RE-GCN \cite{regcn}. The formulation is as follows:
\begin{equation}
    \textbf{o}^{(l+1)}_{t-1}=\sigma(\sum_{(s,r,o)\in \mathcal{F}_{t-1}}\textbf{W}^{(l)}_1(\textbf{s}^{(l)}_{t-1}+\textbf{r}^{(l)}_{t-1})+\textbf{W}^{(l)}_2\textbf{o}^{(l)}_{t-1})\textnormal{,}
\end{equation}
where $\mathcal{F}_{t-1}$ stands for the set of facts within the $t-1$ snapshot, $\textbf{s}_{t-1}, \textbf{o}_{t-1} \in \textbf{E}^{g'}_{t-1}$, $\textbf{r}_{t-1} \in \textbf{R}_{t-1}$, $\mathbf{W}^{(l)}_1$ and $\mathbf{W}^{(l)}_2$ are linear transformations for $(s, r)$ and the self-loop of objects in this layer, $\sigma$ represents the Randomized leaky Rectified Linear Unit (RReLU) activation function. The output of aggregated entity representation intra-snapshot is denoted as $\vec{\textbf{E}}_{t-1}$.
\paragraph{Entity Evolution}
In the temporal aspect, we incorporate a Gated Recurrent Unit (GRU) cell to progressively capture evolving entity representations from each recent snapshot to fit future facts, which can be expressed as:
\begin{equation} \label{ru1}
    \textbf{E}^{g}_t = \mathrm{GRU}(\vec{\textbf{E}}_{t-1},\textbf{E}^{g'}_{t-1})\textnormal{.}
\end{equation}
\paragraph{Relation Updating} \label{sec:ru}
Additionally, HisRES enriches the evolution of relations and simultaneously updates the relation representation in CompGCN during entity aggregation. The operation is as follows:
\begin{equation} \label{ru2}
    \textbf{R}^{(l+1)}_{t-1}=\sigma(\textbf{W}^{(l)}_r\textbf{R}^{(l)}_{t-1})\textnormal{,}
\end{equation}
where $\sigma$ is RReLU activation function and $\textbf{W}^{(l)}_r$ denotes a linear layer. $\vec{\textbf{R}}_{t-1}$ represents the output relation representation of CompGCN. Finally, the relation embeddings $\textbf{R}_t$ are updated similarly to entities:
\begin{equation} 
    \textbf{R}_t = \mathrm{GRU}(\vec{\textbf{R}}_{t-1},pooling(\textbf{E}^{\mathcal{R}}_{t-1}))\textnormal{,}
\end{equation}
where \textit{pooling} denotes the mean pooling operation, $\textbf{E}^{\mathcal{R}}_{t-1}$ denotes the embeddings of $\mathcal{R}$ related entities at $t-1$ snapshot.
\subsubsection{Structural and Temporal Modeling inter-snapshot}
In addition to modeling structural and temporal correlations individually for each nearby snapshot, there are a large number of facts in TKGs that have a sequential correlation. As an example in the ICEWS dataset, (\textit{Citizen} (\textit{Malaysia}), \textit{Make optimistic comment}, \textit{Ministry} (\textit{Malaysia}), \textit{t}) is actually a follow-up event to (\textit{Ministry} (\textit{Malaysia}), \textit{Responses}, \textit{Citizen} (\textit{Malaysia}), \textit{t}-1). In this instance, the prior module designed for each individual snapshot fails to capture correlations such as this from a simple two-hop link, causing a complex mode.

To tackle this issue, we merge every $\omega$ consecutive snapshots into a unified graph, which enables us to capture multi-hop association inter-snapshot using the same aggregation function as the former module (without sharing parameters). The diagram can be seen in the \textit{inter-snapshot} of Figure \ref{mee}. We omit the aggregation step, and denote the output of this process as $\textbf{E}^{gg'}_{t-2:t-1}$. The evolutionary modeling proceeds as follows:
\begin{equation}
    \textbf{E}^{gg}_t = \mathrm{GRU}(\textbf{E}^{gg'}_{t-2:t-1},\textbf{E}^{gg}_{t-2:t-1})\textnormal{.}
\end{equation}
Notably, inter-snapshot modeling involves a sequence of composite graphs spanning larger time intervals with multiple timestamps. Consequently, the time encoding used to represent the time gap relative to the predicted time is not suitable for this module.

\subsection{Self-gating Mechanism} \label{sec:sgm}
HisRES incorporates a self-gating mechanism to adaptively merge multi-granular embeddings from the aforementioned encoder. Given that we capture structure and evolution patterns across snapshots at various granularities, it is crucial for HisRES to effectively combine these features to comprehend correlations among recent snapshots. The self-gating is denoted as $\Theta \in \mathbb{R}^{|\mathcal{E}| \times d}$, which regulates a learnable weighting of different recent entity representations. Formally,
\begin{equation} \label{gate1}
    \textbf{E}_t = \Theta^{\mathcal{E}}\textbf{E}^{g}_t + (1 - \Theta^{\mathcal{E}})\textbf{E}^{gg}_t \textnormal{,}
\end{equation}
\begin{equation} 
    \Theta^{\mathcal{E}} = \sigma(\textbf{W}_3\textbf{E}^{g}_t+\textbf{b}_g){,}
\end{equation}
where $\sigma(\cdot)$ is the \textit{Sigmoid} activation function to limit the weight of $\textbf{E}^{g}_t$ and $\textbf{E}^{gg}_t$ between $[0, 1]$, $\textbf{W}_3\in\mathbb{R}^{d \times d}$ is a linear transformation layer and $\textbf{b}_g$ denotes bias.

\subsection{Global Relevance Encoder} \label{sec:gre}
While the former modules adeptly capture the most recent evolutionary patterns from a multi-granularity perspective, it falls short in highlighting pivotal events closely related to queries within the entire historical TKGs. These events can significantly impact future trends and deserve special consideration. Unlike the widely used method of making statistics from historically related facts \cite{cygnet,tirgn,cenet}, we propose an effective approach to establish a globally relevant graph that amalgamates historically relevant events. This enables explicit modeling of correlations among crucial entities and leverages comprehensive historical interactions. Furthermore, it is worth noting that our strategy, compared to recent methods \cite{hgls,logCL} that construct similar global graphs to utilize historical information, achieves an even more effective capture of global correlations while requiring fewer nodes and connections.
\begin{figure}[htbp]
\centerline{\includegraphics[width=1\linewidth]{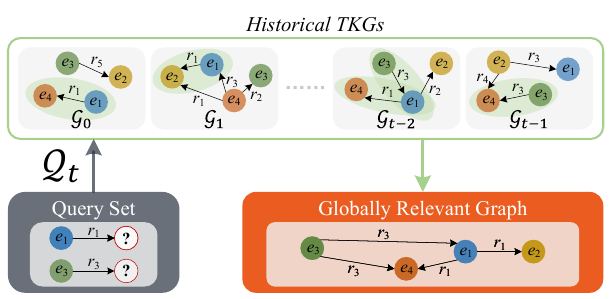}}
\caption{An illustrative construction of the globally relevant graph. In the historical TKGs, query-specific facts are framed by green ovals, while the globally relevant graph is structured by all those facts.}
\label{grs}
\end{figure}
\subsubsection{Global Relevance Structuring}
The idea of structuring global relevance can be seen as an expansion of historical statistics. It involves extracting all related facts $f_q \in {\mathcal{G}_{0:t-1}}$ with query pairs $(s, r)$ from the entire historical TKGs, where $q \in \mathcal{Q}_t$. Subsequently, the globally relevant graph $\mathcal{G}^{\mathcal{H}}_t$ is constructed based on these query-specific facts. This structuring process is illustrated in Figure \ref{grs}. For queries $(e_1, r_1, ?, t)$ and $(e_3, r_3, ?, t)$, HisRES first retrieves relevant facts from all historical snapshots (denoted as ${\mathcal{G}_{0}:\mathcal{G}_{t-1}}$), such as ${(e_1, r_1, e_4)\in\mathcal{G}_{0}}$, ${(e_1, r_1, e_2)\in\mathcal{G}_{1}}$, and ${(e_3, r_3, e_1)\in\mathcal{G}_{t-2}}$. These facts are then integrated into the globally relevant graph.
\subsubsection{ConvGAT}
We consider using the fusion of multi-granularity evolutionary encoder from self-gating as the initial entity embeddings in global relevance aggregation. Since our objective of global relevance encoder is to figure out the important correlations from the whole historically relevant facts, we design a convolution-based graph attention network called ConvGAT, which integrates a 1-D convolutional operator to merge entities and relations, and an attention mechanism to focus on those crucial correlations. 

Firstly, for each link in $\mathcal{G}^{\mathcal{H}}_t$, we calculate the attention scores $\alpha_{o,s}$ based on the embeddings of quadruples, which as follows:
\begin{equation}
    \theta^{(l)}_{o,s} = \frac{exp(\textbf{W}^{(l)}_4\sigma(\textbf{W}^{(l)}_5[\textbf{s}^{(l)}||\textbf{r}||\textbf{o}^{(l)}]))}{\sum_{s' \in \mathcal{N}_{(o)}}exp(\textbf{W}^{(l)}_4\sigma(\textbf{W}^{(l)}_5[\textbf{s}'^{(l)}||\textbf{r}||\textbf{o}^{(l)}])}\textnormal{,}
\end{equation}
where $\textbf{s}, \textbf{o} \in \textbf{E}_t$, $\textbf{r} \in \textbf{R}_t$, $\textbf{W}^{(l)}_4\in\mathbb{R}^{3d}$ and $\textbf{W}^{(l)}_5\in\mathbb{R}^{3d \times 3d}$ represent linear transformation layers, $\sigma$ corresponds to the LeakyReLU activation function, $\mathcal{N}_{(o)}$ denotes the neighboring subject set of object $o$. Notably, we do not update relation embeddings in global relevance encoder. Subsequently, for each link in $\mathcal{G}^{\mathcal{H}}_t$, the aggregation function of global relevance encoder is defined as follows:
\begin{equation}
    \textbf{o}^{(l+1)}=\sigma(\sum_{(s,r,o)\in \mathcal{G}^{\mathcal{H}}_t} \theta^{(l)}_{o,s}\textbf{W}^{(l)}_6\psi(\textbf{s}^{(l)}+\textbf{r})+\textbf{W}^{(l)}_7\textbf{o}^{(l)}) \textnormal{,}
\end{equation}
where $\sigma$ is the RReLU activation function, $\textbf{W}^{(l)}_6$ and $\textbf{W}^{(l)}_7$ are linear transformations, $\psi$ denotes a 1-d convolution operator for the fusion of entity and relation embeddings. The output matrix of all entity embeddings from global relevance encoder can be represented as $\textbf{E}^{\mathcal{H}}_t$.
\subsection{Decoder} \label{dec}
To achieve comprehensive predictions by leveraging entity and relation information from TKGs, HisRES adopts ConvTransE \cite{convtranse} as the score function, a widely used decoder in TKG reasoning, defined as follows:
\begin{equation}
    p(o|s, r, t, \mathcal{G}_{0:t-1}) = softmax(\mathrm{ConvTransE}(\textbf{s}^{\phi},\textbf{r}){\textbf{E}^{\phi}_t})\textnormal{,}
\end{equation}
\begin{equation} \label{gate2}
    \textbf{E}^{\phi}_t = \Theta^{\mathcal{H}}\textbf{E}^{\mathcal{H}}_t + (1 - \Theta^{\mathcal{H}})\textbf{E}_t {,}
\end{equation}
\begin{equation}
    \Theta^{\mathcal{H}} = \sigma(\textbf{W}_8\textbf{E}^{\mathcal{H}}_t+\textbf{b}_{gg}) {,}
\end{equation}
where $\textbf{W}_8$ and $\textbf{b}_{gg}$ denote linear layer and bias, $\textbf{s}^{\phi} \in \textbf{E}^{\phi}_t$, and $\textbf{E}^{\phi}_t$ represents the combined output of self-gating from the multi-granularity evolutionary encoder and global relevance encoder. It is worth noting that the relation prediction is calculated similarly.
\subsection{Learning Objective} \label{lo}
Entity prediction in KGs can be regarded as a multi-classification task. Hence, HisRES employs Cross-Entropy as the loss function. Importantly, we optimize the relation prediction to enhance relation representations, akin to TiRGN \cite{tirgn} and LogCL \cite{logCL}, to further enhance the performance of entity prediction. In addition, a coefficient $\alpha$ is introduced to control the relative weight of the two tasks. The learning objective is formalized as follows:
\begin{equation}
    \begin{aligned}
    \mathcal{L} &= \alpha\sum_{(s,r,t)\in\mathcal{Q}^e_t}y^e_t \log p(o|s, r, t, \mathcal{G}_{0:t-1}) \\
    &+ (1-\alpha)\sum_{(s,o,t)\in\mathcal{Q}^r_t}y^r_t \log p(r|s, o, t, \mathcal{G}_{0:t-1})\textnormal{,}
    \end{aligned}
\end{equation}
where $\mathcal{Q}^e_t$ and $\mathcal{Q}^r_t$ represent the queries of entities and relations at time $t$, and $y^e_t$ and $y^r_t$ denote the truth labels for the two tasks.
\subsection{Complexity Analysis}
We conduct complexity analysis for the proposed HisRES to showcase the efficiency and our trade-offs in capturing multi-granular and global relevant interactions. Note that the embedding dimension $d$ is omitted from the notation. The time complexity of the multi-granularity evolutionary encoder is $O(|2l-\omega+1|(|\mathcal{E}|+|\mathcal{R}|))$, as we consider both intra- and inter-snapshot correlations simultaneously. The time complexity of the global relevance encoder is $O(|\mathcal{E}|+2|\mathcal{F}_{\mathcal{Q}t}|)$, where $|\mathcal{F}_{\mathcal{Q}_t}|$ denotes the number of query-specific facts in $\mathcal{G}^{\mathcal{H}}_t$. Therefore, the total computational complexity can be simplified to $O(|2l-\omega|(|\mathcal{E}|+|\mathcal{R}|)+2|\mathcal{F}_{\mathcal{Q}_t}|)$. This analysis demonstrates that HisRES achieves outstanding performance without a significant increase in computational complexity. The computational complexity of HisRES is compared with baseline methods in Table \ref{complex_ana}. Here, $l$ represents the snapshot length, $\mathcal{M}_{c}$ denotes the convolution operations in TiRGN, while $P$ and $4mC$ correspond to the attention mechanism and contrastive learning components in LogCL, respectively. The additional computational overhead primarily originates from two aspects: the multi-granularity processing (i.e., $|l - \omega|$) and the global relevance structuring (i.e., $|\mathcal{F}_{\mathcal{Q}_t}|$). Although this approach involves comprehensive historical information and intensive correlation analysis, it remains computationally tractable and effectively serves the task objectives.
\begin{table}[ht]
\caption{Computational Complexity Comparison of Baselines with HisRES.}
\centering
\begin{tabular}{ccc}
\hline
Model & Computational Complexity \\
\hline
TiRGN & $\mathcal{O}(|l|(|\mathcal{E}|+|\mathcal{R}|+\mathcal{M}_{c})+|\mathcal{T}|)$\\
LogCL & $O(|l|(|\mathcal{E}|+|\mathcal{R}|+P)+2|\mathcal{F}||\mathcal{T}|+4mC)$ \\ \hline
HisRES & $O(|2l-\omega|(|\mathcal{E}|+|\mathcal{R}|)+2|\mathcal{F}_{\mathcal{Q}_t}|)$ \\
\hline
\end{tabular}%
\label{complex_ana}
\end{table}
\section{Experiments}
\begin{table*}[ht!]
\caption{Statistics of ICEWS14s, ICEWS18, ICEWS05-15 and GDELT datasets.}
\resizebox{\textwidth}{!}{%
\begin{tabular}{cccccccc}
\toprule
Dataset & Entities $\mathcal{E}$ & Relations $\mathcal{R}$ & Training Facts & Validation Facts & Testing Facts  & Timestamps $\mathcal{T}$ & Time Granularity \\
\midrule
ICEWS14s & 7,128 & 230 & 74,845 & 8,514 & 7,371  & 365 & 1 day \\
ICEWS18 & 23,033 & 256 & 373,018 & 45,995 & 49,545  & 304 & 1 day \\
ICEWS05-15 & 10,488 & 251 & 368,868 & 46,302 & 46,159 & 4,017 & 1 day \\
GDELT & 7,691 & 240 & 1,734,399 & 238,765 & 305,241 & 2976 & 15 mins \\
\bottomrule
\end{tabular}%
}
\label{dataset}
\end{table*}
In this section, we perform extensive experiments on four event-based TKG datasets to demonstrate the effectiveness and state-of-the-art performance of our model, which can be summarized in the following questions.
\begin{itemize}[leftmargin=*]
\item \textbf{Q1}: How does HisRES perform compared to state-of-the-art TKG reasoning studies?
\item \textbf{Q2}: How do the multi-granularity evolutionary encoder and global relevance encoder affect the performance of HisRES?
\item \textbf{Q3}: How do the adaptive fusion of different entity representations and the updating of relations affect the performance of HisRES?
\item \textbf{Q4}: How does the proposed superior ConvGAT affect the performance of HisRES?
\item \textbf{Q5}: How does the performance fluctuation of HisRES with different settings?
\end{itemize}
\subsection{Experimental Settings} \label{Exp_settings}
\subsubsection{Datasets}
We evaluate our approach on four real-world event-based TKGs extensively utilized in prior studies. Three datasets are derived from the Integrated Crisis Early Warning System (ICEWS) \cite{icews} with daily temporal granularity: ICEWS14s, ICEWS18, and ICEWS05-15. ICEWS14s and ICEWS18 both cover one-year periods, but ICEWS18 contains approximately three times more entities and five times more facts per snapshot than ICEWS14s. ICEWS05-15 encompasses an extended timespan of 11 years. To demonstrate the broad applicability, scalability, and robustness of HisRES through comprehensive evaluation, we also include the Global Database of Events, Language, and Tone (GDELT) \cite{gdelt}. This dataset features ultra-fine-grained temporal resolution (15-minute snapshots) and high event density (2,278,405 facts within a one-month period). Performance gains on this challenging large-scale dataset provide a more rigorous validation of the model's effectiveness. Consistent with benchmarks \cite{regcn,logCL}, all methods adhere to the split strategy that divides datasets into training (80\%), validation (10\%), and testing (10\%) subsets based on their chronological sequences. Table \ref{dataset} provides additional details on the statistical attributes of these datasets.

Notably, results on some benchmarks such as WIKI and YAGO are not reported. These datasets have large time granularity (one year per snapshot) and long time spans (232 and 189 years, respectively). Such coarse-grained, long-term datasets may be less time-sensitive and less suitable for TKG applications, as knowledge spanning over a century may not be as relevant. Consequently, most recent SOTAs (e.g., L$^2$TKG and LogCL) do not report results on these datasets.
\subsubsection{Baselines}
We compare our proposed HisRES with static KG reasoning methods \cite{distmult,complex,conve,convtranse,rotate,gcn,rgcn,gat} and the state-of-the-art TKG extrapolation models, 
including RE-NET \cite{renet}, CyGNet \cite{cygnet}, xERTE \cite{xerte}, RE-GCN \cite{regcn}, EvoKG \cite{evokg}, CEN \cite{cen}, TiRGN \cite{tirgn}, CENET \cite{cenet}, RETIA \cite{RETIA}, RPC \cite{rpc}, and LogCL \cite{logCL}. HGLS and L$^2$TKG are not reported due to different experimental settings. The details of these extrapolation methods are shown below:
\begin{itemize}[leftmargin=*]
\item RE-NET \cite{renet} utilizes a combination of graph and sequential modeling to capture temporal and structural interactions.
\item CyGNet \cite{cygnet} constructs a historical vocabulary associated with all past events to rectify predictions.
\item xERTE \cite{xerte} proposes a time-aware GAT that leverages temporal relations to explore subgraphs using temporal information.
\item RE-GCN \cite{regcn} incorporates GNN and a recurrent unit to capture both the structural and temporal dependencies of TKGs, and introduces a static graph to augment initial embeddings.
\item CEN \cite{cen} employs a length-aware CNN within a curriculum learning and an online configuration to dynamically accommodate shifts in evolutionary patterns.
\item TiRGN \cite{tirgn} proposes time-aware decoders and incorporates a global historical vocabulary to predict based on whole history.
\item CENET \cite{cenet} introduces contrastive learning into TKGs and designs a classifier to distinguish historical or global events.
\item RETIA \cite{RETIA} combines line graphs and captures the spatial-temporal patterns of entities and relations simultaneously.
\item RPC \cite{rpc} incorporates a rule-based line graph for relations and mines graph structural correlations and periodic temporal interactions via two correspondence units.
\item LogCL \cite{logCL} proposes an entity-aware attention encoder to capture crucial facts from local and global history.
\end{itemize}
\subsubsection{Implementation Details}
The initial embedding dimension $d$ and training epoch are configured to 200 and 30 for all datasets, respectively. The dropout rate of 0.2 is applied uniformly across all modules. The granularity span $\omega$ and hidden layers of GNNs are set to 2 for all datasets. Additionally, we utilize the Adam optimizer with a learning rate parameterized to 0.001. For the optimal historical length $l$ in multi-granularity evolutionary encoder, we conduct grid research according to the validation results, and the best value of $l$ is set to 9, 9, 10 and 7 for ICEWS14s, ICEWS18, ICEWS05-15 and GDELT, respectively. For fair comparison, the task coefficient $\alpha$ is empirically set to 0.7, and the static graph is employed on ICEWS datasets similar to our baselines \cite{regcn,tirgn,rpc,logCL}. All baseline results are reported by previous works. For the global relevance encoder, we adopt a two-phase forward propagation strategy for both raw and inverse query sets, similar to LogCL \cite{logCL}. This approach enables our model to construct globally relevant graphs for the raw and inverse query sets separately, thereby preventing potential data leakage. All experiments are conducted on NVIDIA A800 GPUs.
\subsubsection{Evaluation Metrics}
We report the widely-used time-aware filtered metrics: Mean Reciprocal Rank (MRR) and Hits@\{1,3,10\} \cite{xerte,tirgn,rpc,logCL}, with results presented in percentage form.
\begin{table*}[ht!]
\caption{TKG entity extrapolation results on ICEWS14s, ICEWS18, ICEWS05-15 and GDELT. The time-filtered MRR, H@1, H@3, and H@10 metrics are multiplied by 100. The best results are boldfaced and the second SOTAs are underlined.}
\resizebox{\textwidth}{!}{%
\begin{tabular}{l|cccccccccccccccc}
\toprule
\multirow{2}{*}{Model} & \multicolumn{4}{c}{ICEWS14s} & \multicolumn{4}{c}{ICEWS18} & \multicolumn{4}{c}{ICEWS05-15} & \multicolumn{4}{c}{GDELT} \\ \cline{2-17} 
 & MRR & H@1 & H@3 & H@10 & MRR & H@1 & H@3 & H@10 & MRR & H@1 & H@3 & H@10 & MRR & H@1 & H@3 & H@10 \\ \midrule
DistMult & 15.44 & 10.91 & 17.24 & 23.92 & 11.51 & 7.03 & 12.87 & 20.86 & 17.95 & 13.12 & 20.71 & 29.32 & 8.68 & 5.58 & 9.96 & 17.13 \\
ComplEx & 32.54 & 23.43 & 36.13 & 50.73 & 22.94 & 15.19 & 27.05 & 42.11 & 32.63 & 24.01 & 37.50 & 52.81 & 16.96 & 11.25 & 19.52 & 32.35 \\
ConvE & 35.09 & 25.23 & 39.38 & 54.68 & 24.51 & 16.23 & 29.25 & 44.51 & 33.81 & 24.78 & 39.00 & 54.95 &16.55 &11.02 & 18.88 & 31.60 \\
ConvTransE & 33.80 & 25.40 & 38.54 & 53.99 & 22.11 & 13.94 & 26.44 & 42.28 & 33.03 & 24.15 & 38.07 & 54.32 & 16.20 & 10.85 & 18.38 & 30.86 \\
RotatE & 21.31 & 10.26 & 24.35 & 44.75 & 12.78 & 4.01 & 14.89 & 31.91 & 24.71 & 13.22 & 29.04 & 48.16 & 13.45 & 6.95 & 14.09 & 25.99\\
GCN & 32.15 & 23.07 & 35.53 & 50.63 & 24.99 & 16.04 & 27.86 & 42.75 & 33.67 & 24.08 & 36.99 & 53.67 & 18.16 & 11.38 & 18.99 & 31.35\\
R-GCN & 28.14 & 19.43 & 31.95 & 46.02 & 18.04 & 8.57 & 19.28 & 35.68 & 27.43 & 20.15 & 33.49 & 44.62 & 10.93 & 4.59 & 12.17 & 22.38\\
GAT & 28.54 & 20.54 & 31.12 & 44.21 & 20.69 & 13.39 & 22.75 & 35.07 & 31.33 & 22.96 & 34.12 & 48.01 & 17.92 & 11.22 & 18.83 & 30.87\\
\midrule
RE-NET & 36.93 & 26.83 & 39.51 & 54.78 & 29.78 & 19.73 & 32.55 & 48.46 & 43.67 & 33.55 & 48.83 & 62.72 & 19.55 & 12.38 & 20.80 & 34.00 \\
CyGNet & 35.05 & 25.73 & 39.01 & 53.55 & 27.12 & 17.21 & 30.97 & 46.85 & 40.42 & 29.44 & 46.06 & 61.60 & 20.22 & 12.35 & 21.66 & 35.82 \\
xERTE & 40.02 & 32.06 & 44.63 & 56.17 & 29.31 & 21.03 & 33.51 & 46.48 & 46.62 & 37.84 & 52.31 & 63.92 & 19.45 & 11.92 & 20.84 & 34.18 \\
RE-GCN & 42.39 & 31.96 & 47.62 & 62.29 & 32.62 & 22.39 & 36.79 & 52.68 & 48.03 & 37.33 & 53.90 & 68.51 & 19.69 & 12.46 & 20.93 & 33.81 \\
CEN & 43.34 & 33.18 & 48.49 & 62.58 & 32.66 & 22.55 & 36.81 & 52.50 & - & - & - & -  & 21.16 & 13.43 & 22.71 & 36.38 \\
TiRGN & 44.75 & 34.26 & 50.17 & 65.28 & 33.66 & 23.19 & 37.99 & 54.22 & 50.04 & 39.25 & 56.13 & 70.71 & 21.67 & 13.63 & 23.27 & 37.60\\
CENET & 39.02 & 29.62 & 43.23 & 57.49 & 27.85 & 18.15 & 31.63 & 46.98 & 41.95 & 32.17 & 46.93 & 60.43 &20.23 &12.69 &21.70 &34.92 \\
RETIA & 42.76 & 32.28 & 47.77 & 62.75 & 32.43 & 22.23 & 36.48 & 52.94 & 47.26 & 36.64 & 52.90 & 67.76 &20.12 &12.76 &21.45 &34.49 \\
RPC & - & - & - & - & 34.91 & 24.34 & 38.74 & 55.89 & 51.14 & 39.47 & 57.11 & 71.75 & 22.41 & 14.42 & 24.36 & 38.33\\
LogCL & \underline{48.87} & \underline{37.76} & \underline{54.71} & \underline{70.26} & \underline{35.67} & \underline{24.53} & \underline{40.32} & \underline{57.74} & \underline{57.04} & \underline{46.07} & \underline{63.72} & \underline{77.87} & \underline{23.75} & \underline{14.64} & \underline{25.60} & \underline{42.33} \\
\midrule
HisRES & \textbf{50.48} & \textbf{39.57} & \textbf{56.65} & \textbf{71.09} & \textbf{37.69} & \textbf{26.46} & \textbf{42.75} & \textbf{59.70} & \textbf{59.07} & \textbf{48.62} & \textbf{65.66} & \textbf{78.48} & \textbf{26.58} & \textbf{16.90} & \textbf{29.07} & \textbf{46.31} \\ \textit{improve}$\Delta$
 & 3.29\%$\uparrow$ & 4.79\%$\uparrow$ & 3.55\%$\uparrow$ & 1.18\%$\uparrow$ & 5.66\%$\uparrow$ & 7.87\%$\uparrow$ & 6.03\%$\uparrow$ & 3.39\%$\uparrow$ & 3.56\%$\uparrow$ & 5.54\%$\uparrow$ & 3.04\%$\uparrow$ & 0.78\%$\uparrow$ & 11.92\%$\uparrow$ & 15.44\%$\uparrow$ & 13.55\%$\uparrow$ & 9.40\%$\uparrow$\\ \bottomrule
\end{tabular}%
}
\label{exp1}
\end{table*}
\subsection{Main Results (RQ1)}
The performance of HisRES and advanced baselines under time-aware filtered metrics is reported in Table \ref{exp1}. 

To comprehensively assess the performance of HisRES, we initially compare it with static KG reasoning methods (i.e., DistMult, ComplEx, ConvE, ConvTransE, RotatE, GCN, R-GCN, GAT). As existing TKG baselines consistently outperform these methods across all benchmarks, underscoring the importance of considering temporal information and modeling evolutionary patterns within TKGs. Quantitatively, HisRES achieves more progressive improvements in MRR of 64.30\% over ComplEx (translation-based), 58.24\% over ConvE (convolution-based), and 57.41\% over GCN (GNN-based). Furthermore, to provide valuable insights and compelling evidence addressing Q1, we compare HisRES with the best-performing TKG models. It demonstrates that HisRES is the state-of-the-art method, exhibiting a significant average improvement of 6.11\%, 8.41\%, 6.54\% and 3.69\% in time-filtered MRR, Hits@1, Hits@3 and Hits@10, respectively, across four event-based datasets.

Specifically, CyGNet and CENET focus on statistical analysis of historically repetitive facts but tend to overlook capturing structural interactions among snapshots. In contrast, HisRES not only models the structural and temporal dependencies of recent snapshots, but also incorporates a more comprehensive understanding of the entire history with the proposed global relevance encoder. It is more effective than simply rectifying predicted results.

In comparison to a series of approaches \cite{renet,regcn,cen,tirgn,RETIA,rpc} that mainly focus on modeling  each single snapshots from the most recent history, HisRES additionally considers the multi-granularity recent history. Concretely, we learn the interaction of each adjacent snapshot simultaneously (across time) while utilizing multi-hop links of GNNs to correlate the evolution of events intra-snapshot. Besides, we introduce a novel perspective that exclusively concentrates on relevant events throughout the entire history and further filters important links from them through ConvGAT. Although TiRGN considers global historical information, it still faces challenges similar to CyGNet, as it only employs a simple vector to represent global repetitive facts.

Furthermore, HisRES surpasses LogCL which achieves the second-best results and also provides a global perspective on top of recent snapshot modeling. This outstanding performance stems from multi-granularity and considering a self-gating mechanism to adaptively merge entity representations from different historical periods. Meanwhile, a more efficient approach to handling globally relevant facts is also helpful. In detail, we focus solely on facts directly relevant to the queries, thus reducing the number of historical facts. Additionally, integrating attention calculations into ConvGAT assists in selecting high-impact links or facts.

Notably, previous methods have failed to achieve significant improvements on GDELT due to the finer temporal granularity per snapshot (15 minutes), which makes GDELT more time-sensitive. HisRES provides an effective way of capturing multi-granularity features from both local evolution and global history while achieving a significant improvement of 11.92\%, 15.44\%, 13.55\% and 9.40\% in MRR, Hits@1, Hits@3, and Hits@10 on GDELT. This highlights the ability of HisRES to capture correlations within highly time-sensitive datasets, which have the typical characteristics of TKGs.
\subsection{Ablation Studies (RQ2, RQ3 and RQ4)}
To investigate the impact of different encoders (Q2) and components (Q3) within HisRES, as well as the superiority of proposed ConvGAT (Q4), we conduct a series of ablation studies with ICEWS14s and ICEWS18 as representative datasets. The findings and results are summarized in Table \ref{ablation}.
\subsubsection{The Impact of Different Encoders (RQ2)} To evaluate the indispensable local and global relevance structuring in TKG reasoning, we introduce two variants: HisRES-w/o-$\mathcal{G}$ and HisRES-w/o-$\mathcal{G}^{\mathcal{H}}$, indicating the removal of the multi-granularity evolutionary encoder and the global relevance encoder, respectively. The former highlights the importance of capturing evolutionary patterns from different granularities within the most recent snapshots. Conversely, the latter emphasizes the importance of additionally capturing global dependencies apart from local patterns, which notably enhances TKG reasoning. Additionally, HisRES-w/o-\textit{MG} indicates that our multi-granularity evolutionary encoder only uses intra-snapshot modeling. The resulting deterioration in scores demonstrates the importance of considering both intra- and inter-snapshot information simultaneously.
\begin{table}[ht]
\centering
\caption{Ablation studies with time-filtered metrics on ICEWS14s and ICEWS18.}
\resizebox{1\linewidth}{!}{%
\begin{tabular}{l|cccccccc}
\toprule
\multirow{2}{*}{Variant Model} & \multicolumn{4}{c}{ICEWS14s} & \multicolumn{4}{c}{ICEWS18} \\ \cline{2-9}
& MRR & H@1 & H@3 & H@10 & MRR & H@1 & H@3 & H@10 \\
\midrule
HisRES &\textbf{50.48} & \textbf{39.57} & \textbf{56.65} & \textbf{71.09} & \textbf{37.69} & \textbf{26.46} & \textbf{42.75} & \textbf{59.70} \\
\midrule
HisRES-w/o-$\mathcal{G}$ & 45.48 & 34.76 & 50.94 & 65.72 & 29.16 & 18.45 & 33.17 & 50.61 \\
HisRES-w/o-$\mathcal{G}^{\mathcal{H}}$ & 41.83 & 31.49 & 47.01 & 61.74 & 31.55 & 21.53 & 35.41 & 51.48\\
HisRES-w/o-\textit{MG} & 49.67 & 38.95 & 55.55 & 70.11 & 36.31 & 25.11 & 41.09 & 58.49\\
\midrule
HisRES-w/o-$\textit{SG}^1$ & 50.04 & 39.34 & 55.86 & 70.28 & 37.08 & 25.76 & 42.07 & 59.39 \\
HisRES-w/o-$\textit{SG}^2$ & 50.10 & 39.42 & 56.24 & 70.07 & 36.99 & 25.70 & 41.95 & 59.39\\
HisRES-w/o-\textit{TE} & 50.08 & 39.47 & 55.83 & 70.54 & 37.25 & 26.10 & 42.14 & 59.24 \\
HisRES-w/o-\textit{RU} & 50.17 & 39.37 & 56.17 & 70.38 & 36.99 & 25.79 & 41.79 & 59.12 \\
\midrule
HisRES-w/-CompGCN & 48.75 & 37.71 & 54.70 & 69.73 & 36.37 & 25.34 & 41.06 & 58.21 \\
HisRES-w/-RGAT & 47.99 & 36.95 & 53.94 & 69.18 & 35.68 & 24.58 & 40.30 & 57.72 \\
\bottomrule
\end{tabular}%
}
\label{ablation}
\end{table}
\subsubsection{The Impact of Different Components (RQ3)} To investigate the effectiveness of adaptively merging different entity representations, we replace self-gating at two parts (corresponding to equations (\ref{gate1}) and (\ref{gate2})) with simple summation. These variants are denoted as HisRES-w/o-$\textit{SG}^1$ and HisRES-w/o-$\textit{SG}^2$. The decreased performance shown in the second part of Table \ref{ablation} indicates that the self-gating mechanism is effective for either type of entity representation fusion. Subsequently, HisRES-w/o-\textit{TE} (corresponding to equations (\ref{te1}) and (\ref{te2})) exhibits a 0.98\% decrease in MRR, highlighting the impact of considering the temporal gap in our multi-granularity evolutionary encoder. Additionally, compared to HisRES-w/o-\textit{RU} (corresponding to equations (\ref{ru1}) and (\ref{ru2})), our strategy of relation updating achieves an average improvement of 1.26\% in MRR, underscoring the importance of considering the evolution of relations during aggregations.
\subsubsection{The Impact of Different GNNs (RQ4)} To assess the superiority of the proposed ConvGAT, we replace it with two widely used GNN aggregators, CompGCN \cite{compgcn} and RGAT \cite{KBGAT}. The resulting variants are denoted as HisRES-w/-CompGCN and HisRES-w/-RGAT, respectively, as shown in the third part of Table \ref{ablation}. Compared to the second-best result, HisRES achieves an average improvement of 3.59\%, 4.68\%, 3.84\% and 2.26\% in MRR, Hits@1, Hits@3 and Hits@10, respectively. It illustrates the significance of assigning different degrees of attention to various historical facts and underscores the efficiency of ConvGAT in calculating attention scores. Moreover, to demonstrate ConvGAT's ability to focus on crucial historical cues, we conduct an additional analysis of attention weight distributions, as detailed in Section \ref{cs_convgat}.
\subsection{Sensitivity Analysis (RQ5)}
\begin{figure}[htbp]
\centering
\subfloat[Analysis on ICEWS14s.]{
		\includegraphics[width=0.49\linewidth]{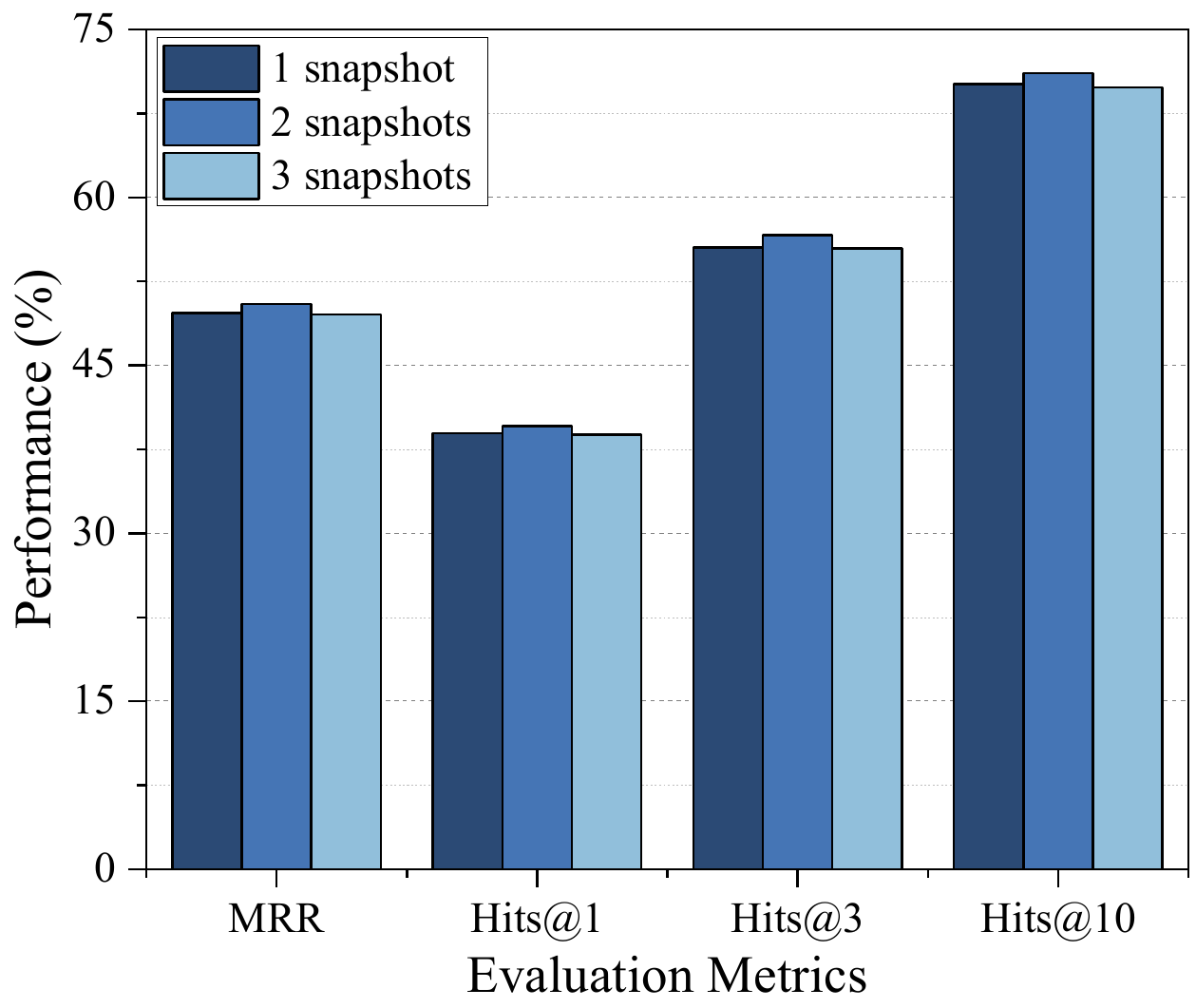}}
\subfloat[Analysis on ICEWS18.]{
		\includegraphics[width=0.49\linewidth]{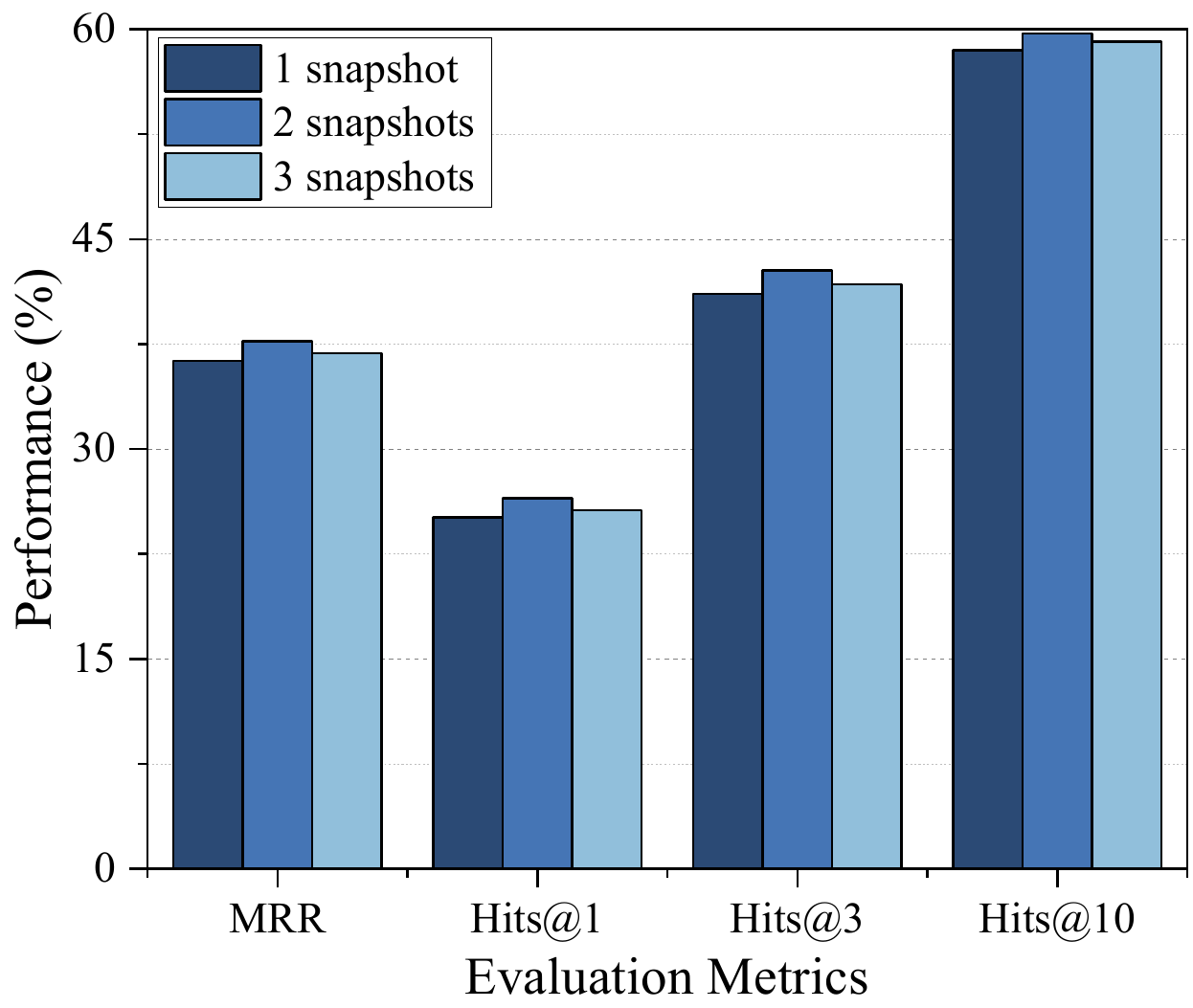}}
\caption{The influence of granularity spans from each single snapshot to every three adjacent snapshots.}
\label{level}
\end{figure}

\begin{figure}[htbp]
\centering
\subfloat[Analysis on ICEWS14s.]{
		\includegraphics[width=0.49\linewidth]{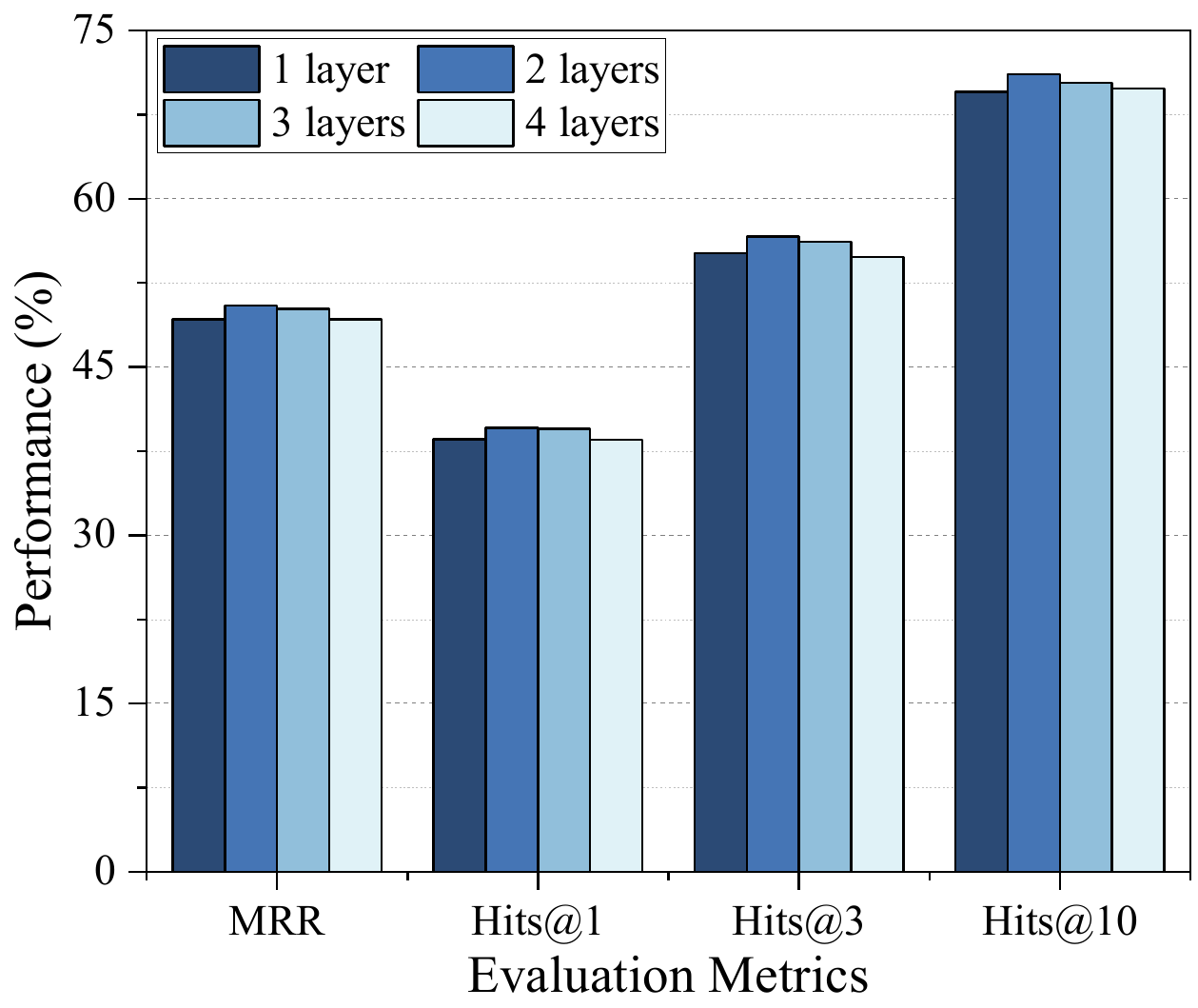}}
\subfloat[Analysis on ICEWS18.]{
		\includegraphics[width=0.49\linewidth]{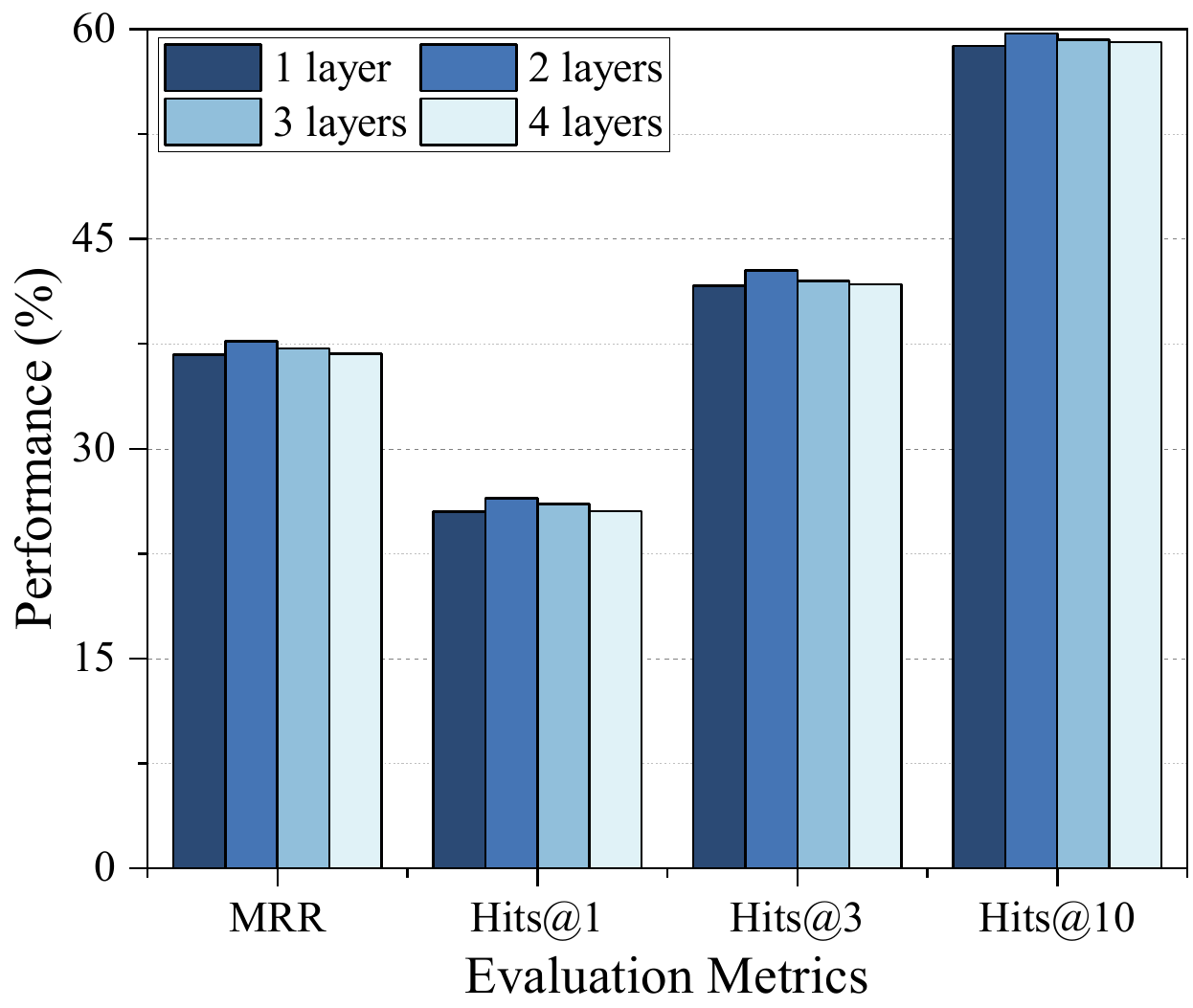}}
\caption{The influence of hidden layers, varying from one layer to four layers of ConvGAT.}
\label{layer}
\end{figure}

To further investigate the sensitivity of HisRES in different settings (Q5), we conduct experiments to explore the influence of granularity span (\S \ref{igs}), hidden layers of GNNs (\S \ref{ihlg}), and historical length (\S \ref{ihl}) on the performance of HisRES, which is shown in Figure \ref{level}, \ref{layer}, and \ref{length}, respectively.
\vspace{-12pt}
\vspace{-2pt}
\subsubsection{Exploring the Influence of Granularity Span} \label{igs}
The granularity span $\omega$ plays a pivotal role in controlling the interaction range inter-snapshot in the proposed multi-granularity evolutionary encoder simultaneously. It determines the quality of entity representations studied from the most recent snapshots. As illustrated in Figure \ref{level}, HisRES exhibits a compelling performance with capturing correlations at each 2 adjacent snapshots level. These findings highlight the impressive ability of HisRES to adapt to historical information from varying time spans. It is worth noting that HisRES maintains strong performance with different levels of granularity, demonstrating the robustness and effectiveness of our basic architecture.
\vspace{-2pt}
\vspace{-2pt}
\subsubsection{Exploring the Influence of Hidden Layer of GNNs} \label{ihlg}
Since HisRES integrates CompGCN and ConvGAT to learn multi-granularity evolutionary and globally relevant patterns, respectively, it is essential to evaluate the hidden layer numbers by investigating the influence of aggregation multi-hop neighbor information on entity representation. The results are depicted in Figure \ref{layer}. Specifically, the two-hop results outperform both the one-hop and three-hop results. This observation suggests that overly simplistic neighborhoods or increasing the number of hops does not significantly enhance the performance of aggregating relevant events. Conversely, it may lead to oversmoothing or introduce unnecessary features that impede effective representations.
\begin{figure}[htbp]
\centering
\subfloat[Analysis on ICEWS14s.]{
		\includegraphics[width=0.49\linewidth]{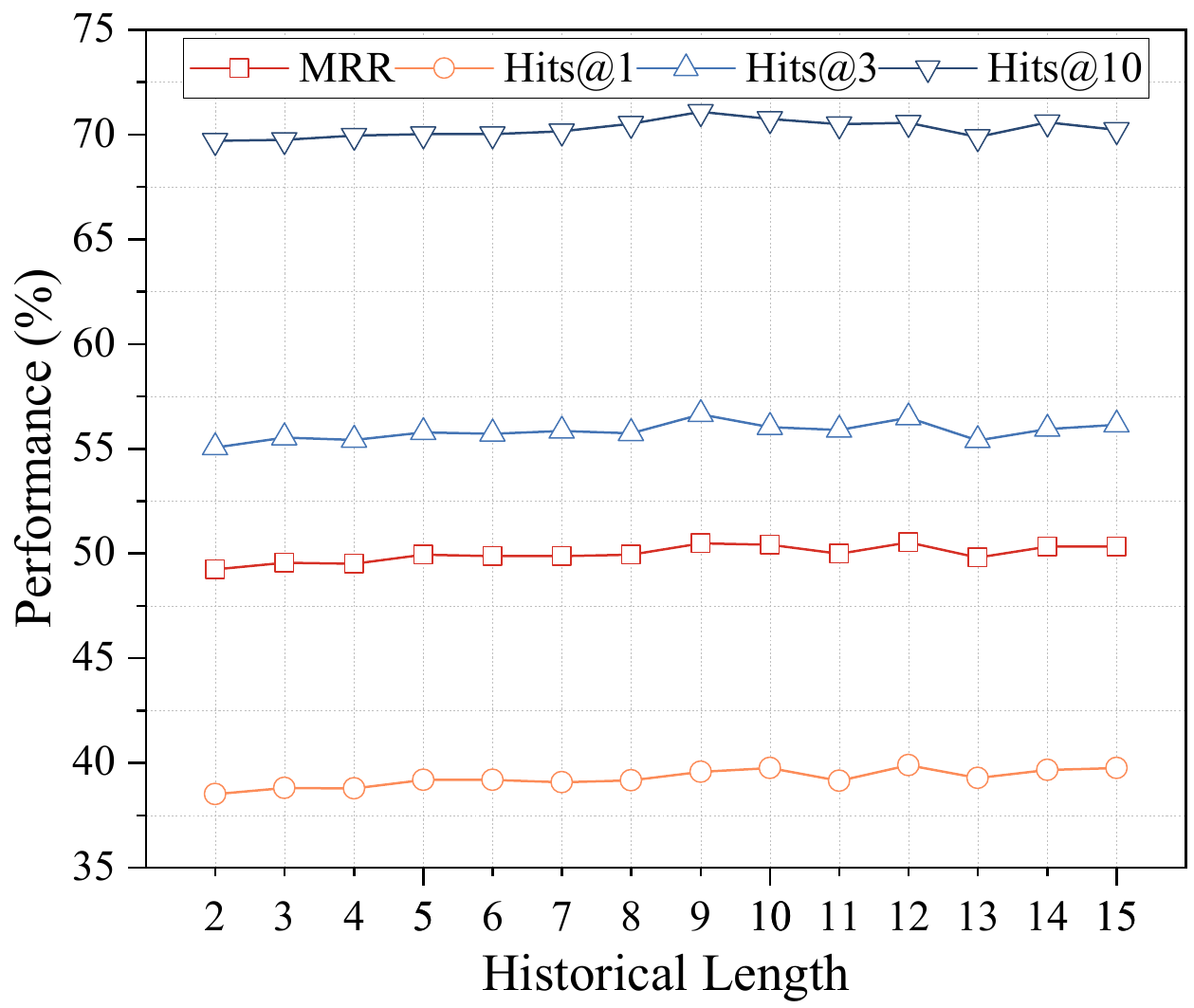}}
\subfloat[Analysis on ICEWS18.]{
		\includegraphics[width=0.49\linewidth]{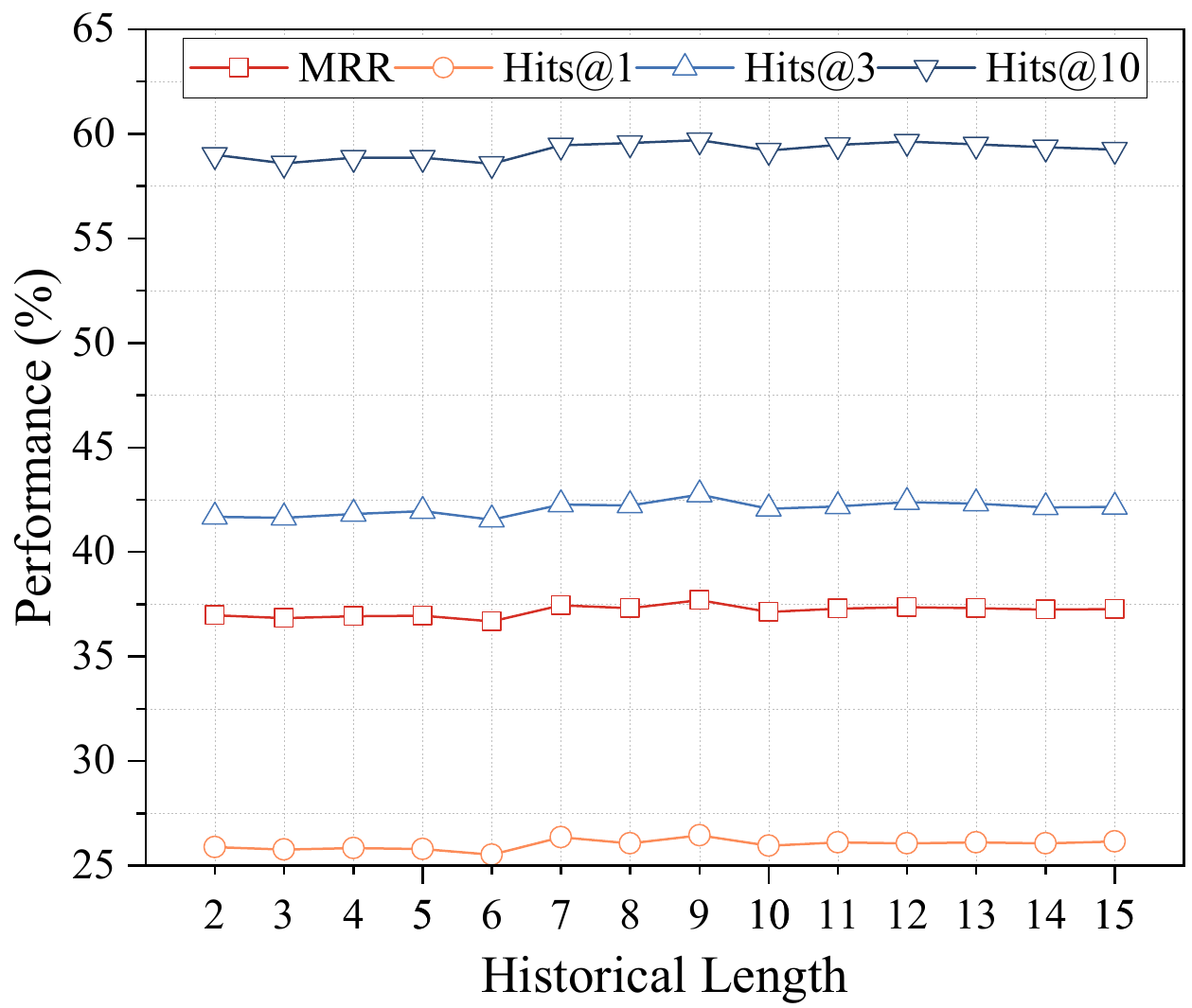}}\\
\subfloat[Analysis on ICEWS05-15.]{
            \includegraphics[width=0.49\linewidth]{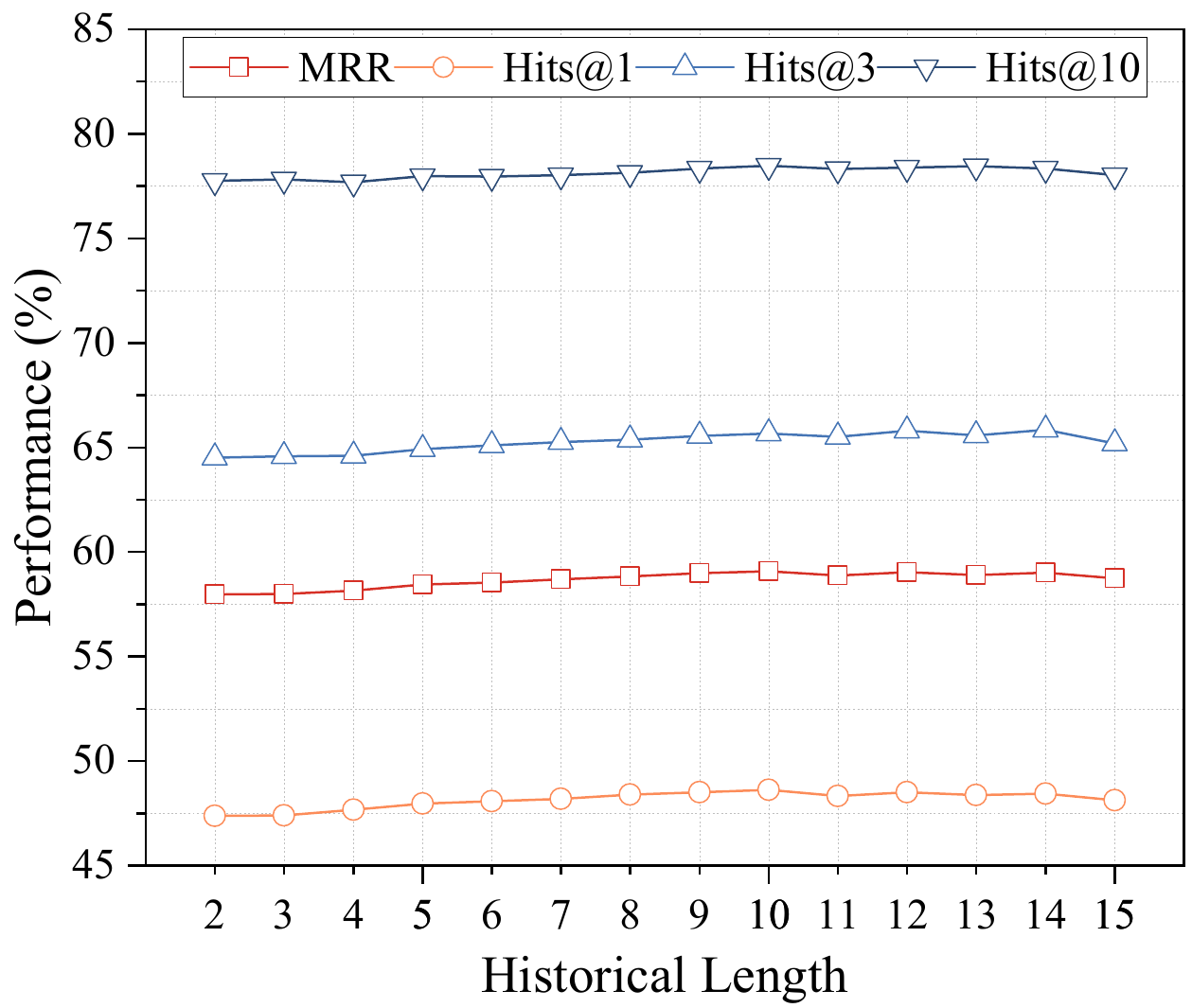}}
\subfloat[Analysis on GDELT.]{
            \includegraphics[width=0.49\linewidth]{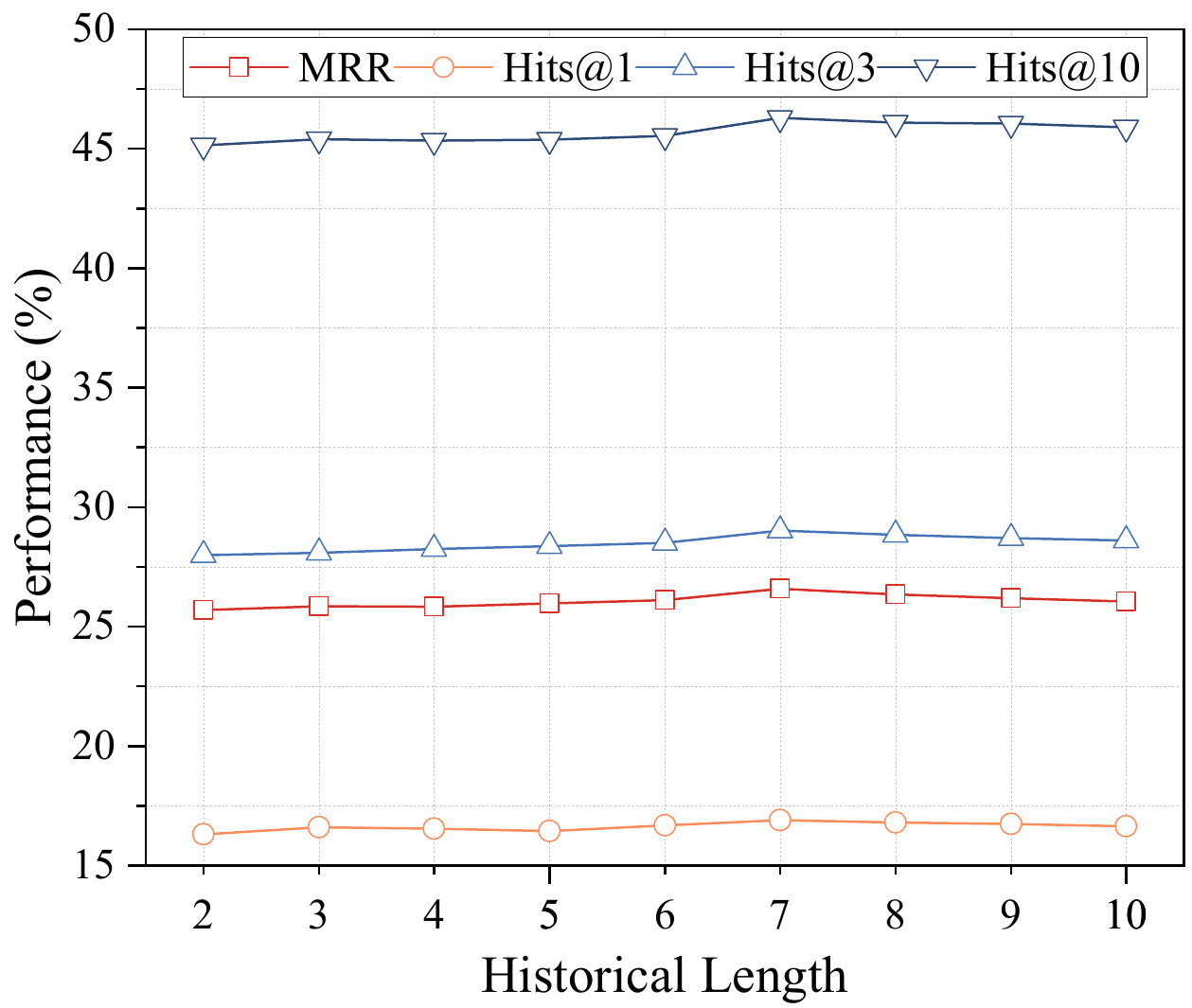}}
\caption{The influence of historical length.}
\label{length}
\end{figure}
\vspace{-6pt}
\subsubsection{Exploring the Influence of Historical Length} \label{ihl}
The historical length $l$ determines the dependency degree of HisRES on recent history. Specifically, this value directly affects the performance of the multi-granularity evolutionary encoder and is expected to be smaller, as we already employ global relevance encoder to learn a wider range of historical knowledge. As illustrated in Figure \ref{length}, when the length of the adjacent history increases, the performance of HisRES exhibits less fluctuation. This observation further demonstrates the effectiveness of HisRES in considering historical information from both local and global perspectives.
\begin{table*}[t!]
\centering
\caption{Detailed execution times of HisRES on ICEWS14s dataset. All times are reported in seconds per epoch.}
\begin{tabular}{l|cccc}
\toprule
Main Modules                                            & Components                               & Details     & Training Time & Inference Time \\ \midrule
\multirow{5}{*}{Multi-granularity Evolutionary Encoder} & \multirow{2}{*}{Modeling intra-snapshot} & Structuring & 7.99          & 0.74           \\
                                                        &                                          & Aggregation & 24.22         & 1.98           \\ \cline{2-5} 
                                                        & \multirow{2}{*}{Modeling inter-snapshot} & Structuring & 8.07          & 0.76           \\
                                                        &                                          & Aggregation & 22.77         & 1.85           \\ \cline{2-5} 
                                                        & Self-gating$^1$                              & -           & 0.04          & $<$0.01           \\ \midrule
\multirow{2}{*}{Global Relevance Encoder}               & Global Relevance Structuring             & -           & 96.04         & 10.42          \\
                                                        & ConvGAT                                  & -           & 28.76         & 4.49           \\ \midrule
Self-gating$^2$                                             & -                                        & -           & 0.06          & $<$0.01           \\ \bottomrule
\end{tabular}
\label{exectime}
\end{table*}
\subsection{Evaluation on the Impact of Self-gating}
To investigate the impact of self-gating in HisRES, we conduct weight analysis to examine how it adaptively merges: 1) multi-granular representations, and 2) representations from the multi-granularity evolutionary encoder and global relevance encoder. Notably, the values for each entity represent the mean across dimension $d$, and the values in 2) are further averaged from both raw and inverse queries.
\begin{figure}[htbp]
\centerline{\includegraphics[width=1\linewidth]{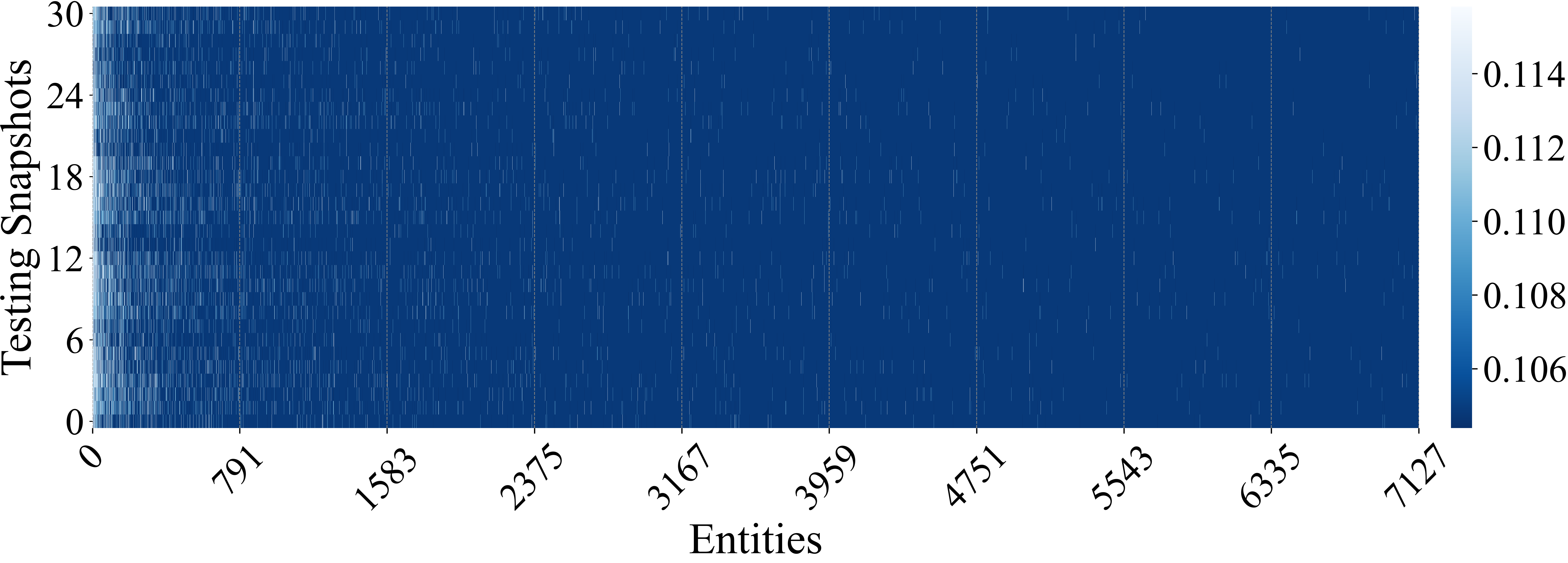}}
\caption{Distribution of self-gating weights between intra- and inter-snapshot encoding on ICEWS14s. Lighter (darker) blue indicates higher (lower) intra-snapshot encoding weights.}
\label{eva-sg1}
\end{figure}
\subsubsection{Weight between intra- and inter-snapshot}
As illustrated in Figure \ref{eva-sg1}, in the first self-gating, most snapshots heavily rely on features modeled by inter-snapshot encoding during testing. The average weight assigned to intra-snapshot modeling is only 0.11. This observation demonstrates the crucial role of historical information from inter-snapshot modeling in TKG reasoning.
\begin{figure}[htbp]
\centerline{\includegraphics[width=1\linewidth]{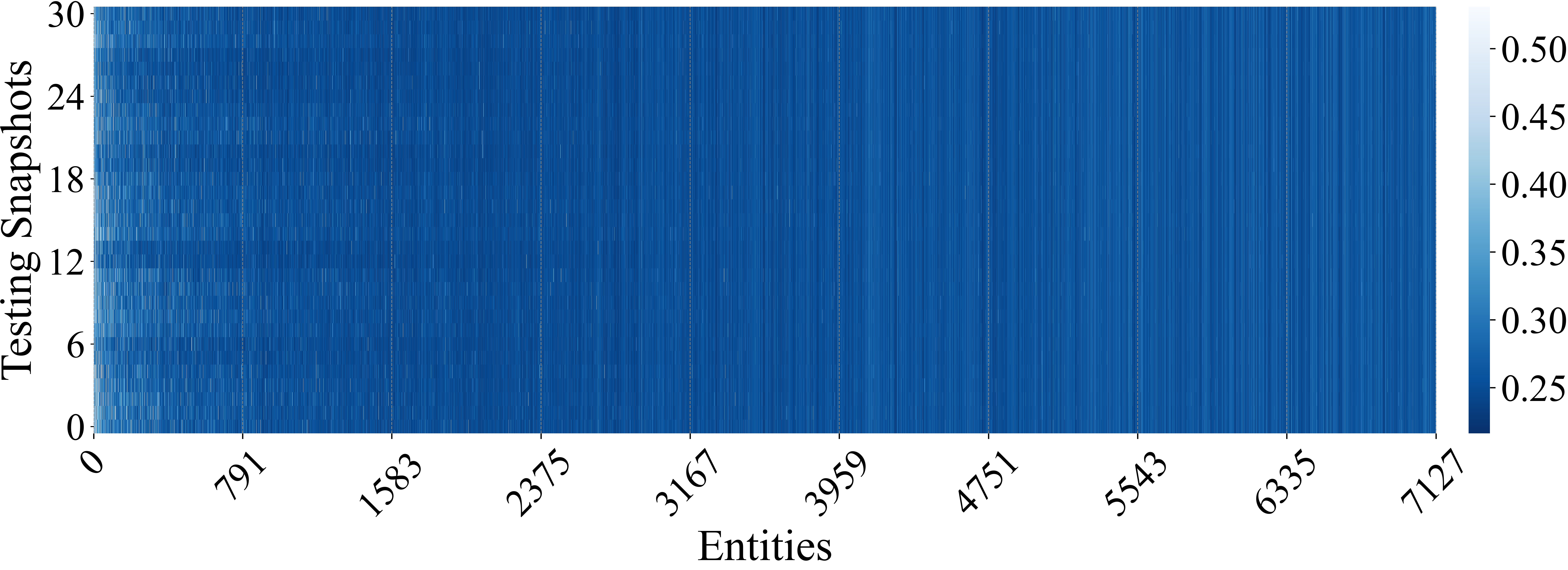}}
\caption{Distribution of self-gating weights between multi-granular and global encoding on ICEWS14s. Lighter (darker) blue indicates higher (lower) multi-granular encoding weights.}
\label{eva-sg2}
\end{figure}
\subsubsection{Weight between Multi-granular and Global History}
As shown in Figure \ref{eva-sg2}, HisRES adaptively balances multi-granularity and global representations through the second self-gating. The average weight assigned to multi-granularity modeling is 0.26, indicating that globally relevant historical information provides numerous effective cues. Consequently, both 1) and 2) demonstrate the effectiveness of self-gating in HisRES. Despite varying degrees of dependence on historical information across entities at different granularities, our model effectively learns appropriate representations from such diverse temporal granularities.
\subsection{Execution Time Analysis}
\begin{figure}[htbp]
\centering
\subfloat[Training time comparison on four datasets.]{
		\includegraphics[width=0.95\linewidth]{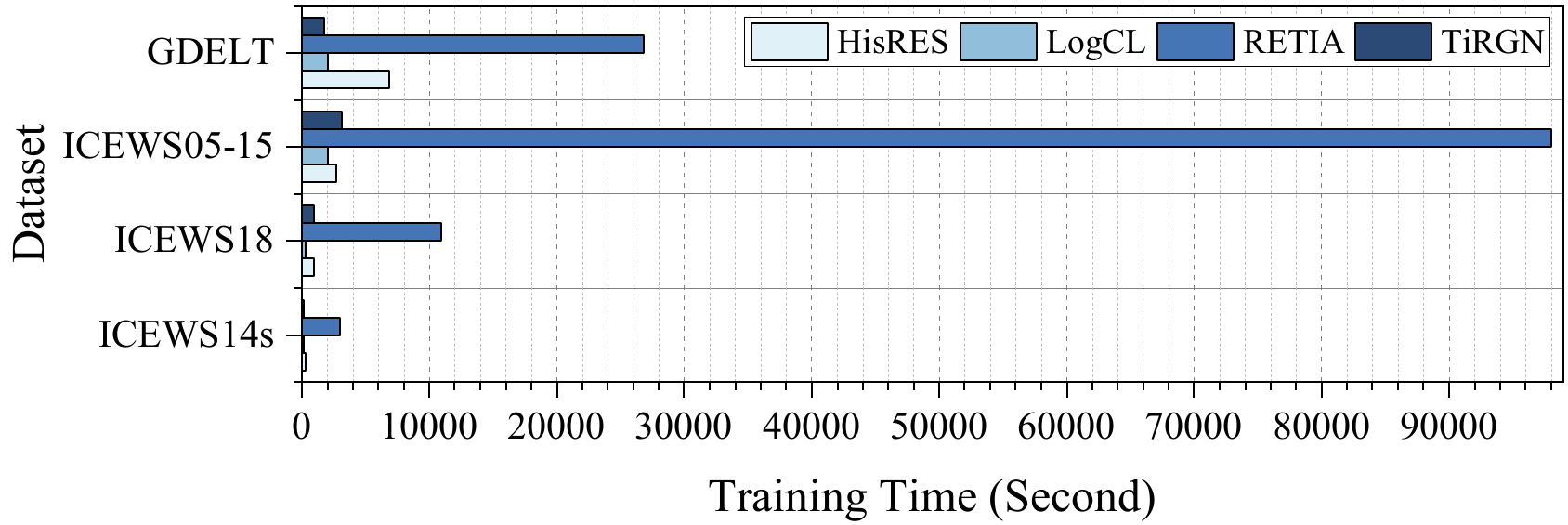}}
\\
\subfloat[Inference time comparison on four datasets.]{
		\includegraphics[width=0.95\linewidth]{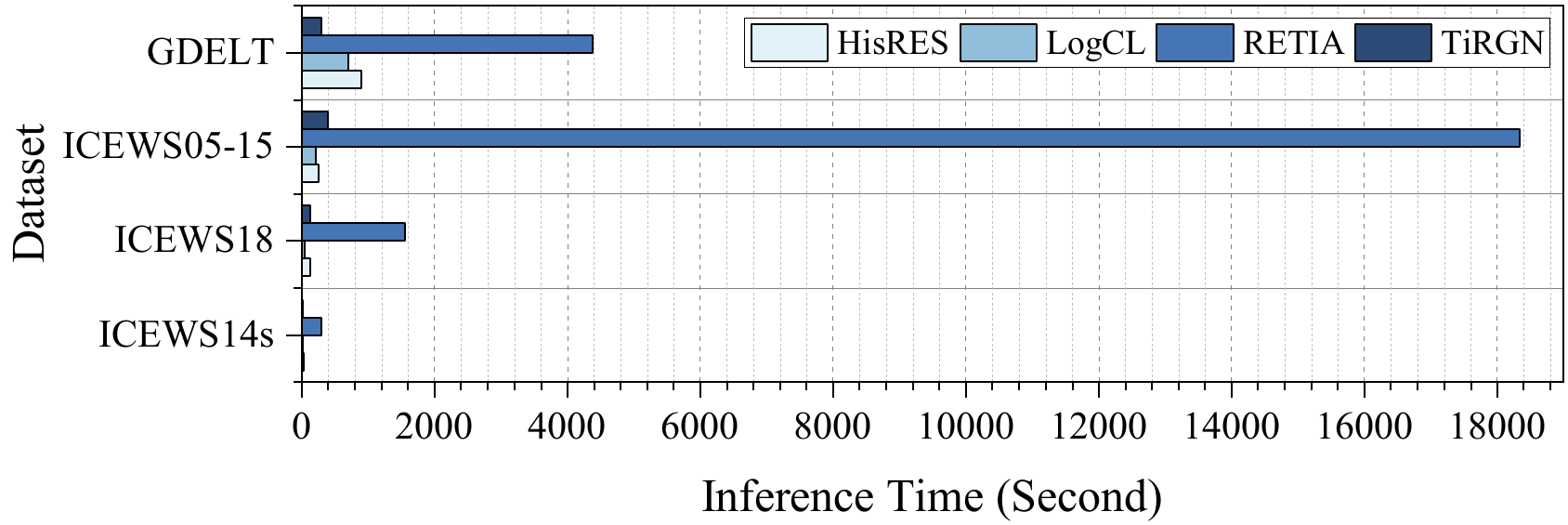}}
\caption{Execution time (seconds per epoch) comparison to baselines on four datasets.}
\label{time}
\end{figure}

To evaluate the computational efficiency and performance trade-offs of HisRES, comprehensive timing analyses are conducted comparing our approach with SOTAs across diverse datasets, as illustrated in Figure \ref{time}. The datasets include ICEWS18 (higher entity count), ICEWS05-15 (extended time span), and GDELT (fine-grained timestamp and fact density), demonstrating the scalability of HisRES. Training and inference times are measured in seconds for consistent comparison. The observed variations in execution time can be attributed to three primary factors:
\begin{enumerate}[leftmargin=*]
\item Variable optimal historical lengths significantly impact local encoding efficiency
\item Global information processing requires comprehensive fact consideration, especially the density of history, which may proportionally affect the efficiency
\item RETIA's line graph construction introduces substantial computational overhead
\end{enumerate}
The results demonstrate that HisRES maintains comparable computational efficiency. The predominant source of computational overhead in HisRES stems from global relevance structuring across the entire historical timeline, which is particularly sensitive to historical correlation density. For instance, ICEWS18 and GDELT exhibit high computational demands due to their dense fact distributions in global graphs. Conversely, ICEWS05-15, despite its large fact count, demonstrates lower density due to its extended temporal span. Notably, direct performance comparisons warrant careful interpretation due to methodological differences in data preprocessing. While existing approaches typically pre-compute and store historical graphs prior to training, HisRES dynamically generates multi-granular and globally relevant graphs during both training and inference phases. Although this real-time computation introduces additional overhead, it significantly enhances the model's flexibility and practical applicability.

All experiments are conducted using NVIDIA A800 80G GPUs, Intel Xeon Silver 4316 CPUs, and 256G RAM. Detailed analyses for the running time of each proposed module of HisRES are presented in Table \ref{exectime}. In the multi-granularity evolutionary encoder, the primary computational overhead stems from two processes: (1) structuring subgraphs at each granularity level (single snapshot or adjacent snapshots) and (2) aggregating information for each subgraph. Our experimental results demonstrate that structuring adjacent snapshots introduces only minimal additional processing time compared to structuring single snapshot subgraphs (8.07 seconds versus 7.99 seconds), with trivial differences in aggregation time (24.22 seconds versus 22.77 seconds). Although our computational analysis indicates that global relevance structuring constitutes a significant portion of HisRES's total overhead, the efficiency of the proposed ConvGAT in processing numerous global historical facts is comparable to the aggregation time required for a single snapshot. Additionally, our results confirm that the self-gating mechanism represents a highly cost-effective module, efficiently balancing weights across different modules with an inference time of less than 0.01 seconds.
\begin{table*}[t]
\caption{Case study of HisRES using three queries from the ICEWS14s dataset, dated 2014-12-01. The ground truth entity is shown in bold. \dag\ denotes attention scores derived from non-query relations. - indicates the objects are not in $\mathcal{G}^{\mathcal{H}}_{2014-12-01}$.}
\centering
\begin{tabular}{llcc}
\toprule
\multicolumn{1}{l}{Query} & Hits@5 of HisRES & Attention Score from ConvGAT & Truth in History\\ \midrule
\multirow{5}{*}{\begin{tabular}[c]{@{}l@{}}\textit{s}: \textit{Philippines} (\textit{a})\\ \textit{r}: \textit{Arrest,\_detain,\_or\_charge\_with\_legal\_action}\\ \textit{t}: 2014-12-01 \\ Number of links in $\mathcal{G}^{\mathcal{H}}_{2014-12-01}$: 47 \end{tabular}}
& \textit{Men\_}\textit{(Philippines)} & 0.0030 & 2014-01-17 \\ & {\textbf{\textit{Criminal\_}(\textit{Philippines})}} (\textit{d})& \textbf{0.0307} & 2014-02-06 \\ & \textit{Abu\_Sayyaf} & 0.0026 & 2014-05-30 \\ & \textit{Employee\_}\textit{(Philippines)} & 0.0025 & 2014-08-27 \\ & \textit{Businessperson\_}\textit{(Philippines)} & 0.0020 & ...... \\\midrule
\multirow{5}{*}{\begin{tabular}[c]{@{}l@{}}\textit{s}: \textit{Police\_}(\textit{Philippines}) (\textit{b})\\ \textit{r}: \textit{Confiscate\_property}\\  \textit{t}: 2014-12-01\\ Number of links in $\mathcal{G}^{\mathcal{H}}_{2014-12-01}$: 8 \end{tabular}}
& {\textbf{\textit{Criminal\_}(\textit{Philippines})}} (\textit{d})& \textbf{0.1126} & 2014-02-21 \\ & \textit{Philippines} & 0.3459\dag & 2014-05-22 \\ & \textit{Abu\_Sayyaf} & 0.0790 & 2014-06-24 \\ & \textit{Men\_}\textit{(Philippines)} & - & 2014-07-13 \\ & \textit{Employee\_}\textit{(Philippines)} & - & ......\\ \midrule
\multirow{5}{*}{\begin{tabular}[c]{@{}l@{}}\textit{s}: \textit{City\_Mayor\_}(\textit{Philippines}) (\textit{c})\\\textit{r}: \textit{Make\_statement}\\ \textit{t}: 2014-12-01\\ Number of links in $\mathcal{G}^{\mathcal{H}}_{2014-12-01}$: 16 \end{tabular}} 
& \textit{Philippines} & 0.3632\dag \\ & \textit{City\_Mayor\_}(\textit{Philippines}) & - & 2014-04-09 \\ & \textit{Police\_}\textit{(Philippines)} & 0.0933 & 2014-09-11 \\ & \textit{Military\_}\textit{(Philippines)} & 0.0261 & 2014-09-18 \\ & \textbf{\textit{Criminal\_}\textit{(Philippines)}} (\textit{d}) & \textbf{0.0957} \\ \bottomrule
\end{tabular}
\label{case_table}
\end{table*}
\subsection{Robustness Analysis}
\begin{figure}[h!]
\centering
\subfloat[MRR Results on ICEWS14s.]{
		\includegraphics[width=0.49\linewidth]{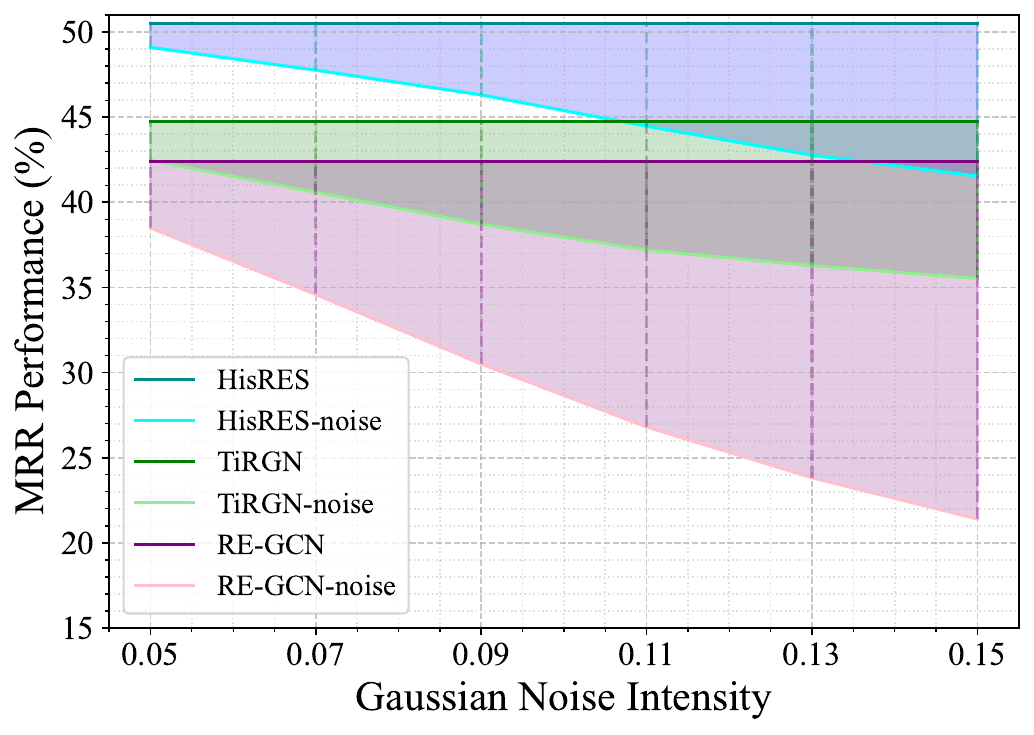}}
\subfloat[MRR Results on ICEWS18.]{
		\includegraphics[width=0.495\linewidth]{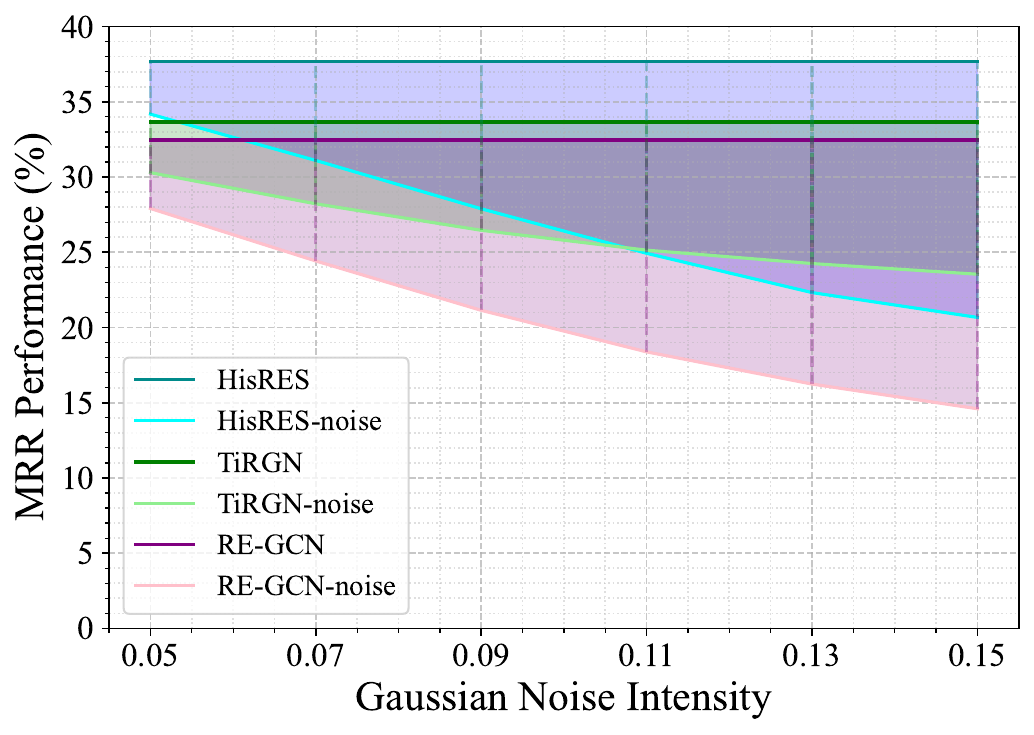}}
\caption{Analysis of MRR performance under varying Gaussian noise intensities on ICEWS14s and ICEWS18. Light blue, green, and purple shading indicate the variation range of HisRES, TiRGN, and RE-GCN, respectively.}
\label{robust}
\end{figure}

To investigate the robustness of HisRES, we evaluated performance by introducing random Gaussian noise of varying intensities to the initial entity embeddings during testing. This evaluation is conducted on HisRES, TiRGN \cite{tirgn}, and RE-GCN \cite{regcn}, with results denoted by the "-noise" suffix in Figure \ref{robust}. Experimental results demonstrate that HisRES maintains robust performance under various noise intensities. This resilience stems from HisRES's global relevance encoder, which effectively identifies highly correlated facts in the global history. Quantitatively, on ICEWS14s, HisRES shows moderate noise resistance with a 17.79\% performance degradation at noise intensity 0.15, compared to degradations of 20.60\% and 49.49\% for TiRGN and RE-GCN, respectively. Though TiRGN's historical information mask matrix helps mitigate performance decline under noise, HisRES maintains better performance for noise intensities below 0.11 on ICEWS18.
\subsection{Case Study of the proposed ConvGAT} \label{cs_convgat}
To demonstrate the capability of HisRES to focus on crucial events within a global historical context and to showcase the effectiveness of our proposed ConvGAT, we select three queries pertaining to \textit{Philippines} from ICEWS14s, all dated 2014-12-01, as illustrative examples:
\begin{enumerate}[leftmargin=*]
    \item (\textit{Philippines}, \textit{Arrest,\_detain,\_or\_charge\_with\_legal\_action}, ? , 2014-12-01);
    \item (\textit{Police\_}(\textit{Philippines}), \textit{Confiscate\_property}, ? , 2014-12-01);
    \item (\textit{City\_Mayor\_}(\textit{Philippines}), \textit{Make\_statement}, ? , 2014-12-01).
\end{enumerate}
All three queries share the ground truth entity \textit{Criminal\_}(\textit{Philippines}). Table \ref{case_table} presents comprehensive information for each query, including: the number of incoming edges for each query's subject in the globally relevant graph $\mathcal{G}^{\mathcal{H}}_{2014-12-01}$, the Hits@5 objects predicted by HisRES, the attention scores calculated by ConvGAT for predicted objects in $\mathcal{G}^{\mathcal{H}}_{2014-12-01}$, and the historical timestamps when the ground truth facts occurred. For readability, we denote the subjects and ground truth as \textit{a} to \textit{d}. These examples, derived from historical events, exhibit varying densities of association facts in $\mathcal{G}^{\mathcal{H}}_{2014-12-01}$ (excluding inverse queries). ConvGAT demonstrates its ability to identify crucial facts from $\mathcal{G}^{\mathcal{H}}_{2014-12-01}$, regardless of their temporal distance, and to effectively aggregate information from entities highly related to \textit{Philippines}.

Table \ref{case_table} reveals that, for all three queries, ConvGAT assigns the highest attention scores to the ground truth \textit{d} under the respective query relations. Specifically, in example 1), despite 47 entities being directly relevant to \textit{a}, HisRES successfully identifies and prioritizes important entities with high predicted scores, even when they are not from the most recent events. As shown in the first part of Table \ref{case_table}, \textit{d} contributes the highest score of 0.0307. For example 2), \textit{d} provides the highest attention to \textit{b}, with a score of 0.1126. Furthermore, ConvGAT effectively aggregates related entities (e.g., \textit{a}) under other relations with a cumulative attention score of 0.3459, demonstrating robust semantic relevance. In example 3), although the ground truth is ranked at a lower position, ConvGAT nonetheless ensures that its global attention score relative to query remains high.
\section{Conclusion}
This paper introduces HisRES, an innovative model for TKG reasoning. It comprises two novel encoders that excel at learning local and global representations by structuring historically relevant events. Specifically, we propose a multi-granularity evolutionary encoder to capture structural and temporal interactions from intra-snapshot and inter-snapshot perspectives. In particular, we introduce a global relevance encoder featuring a distinctive ConvGAT to attend to crucial dependencies throughout the entire history. Furthermore, HisRES incorporates a self-gating mechanism for comprehensive fusion of TKG information. Experimental results on four event-based TKG datasets and various ablation studies demonstrate the effectiveness and superior performance of HisRES. Future research avenues include exploring pruning techniques for global relevance and enhancing the modeling of temporal information.
\section*{Acknowledgments}
This research work is partly supported by National Natural Science Foundation of China [No. 62376055].
\bibliographystyle{IEEEtran}
\bibliography{mybibfile}

\begin{thebibliography}{10}
\providecommand{\url}[1]{#1}
\csname url@samestyle\endcsname
\providecommand{\newblock}{\relax}
\providecommand{\bibinfo}[2]{#2}
\providecommand{\BIBentrySTDinterwordspacing}{\spaceskip=0pt\relax}
\providecommand{\BIBentryALTinterwordstretchfactor}{4}
\providecommand{\BIBentryALTinterwordspacing}{\spaceskip=\fontdimen2\font plus
\BIBentryALTinterwordstretchfactor\fontdimen3\font minus \fontdimen4\font\relax}
\providecommand{\BIBforeignlanguage}[2]{{%
\expandafter\ifx\csname l@#1\endcsname\relax
\typeout{** WARNING: IEEEtran.bst: No hyphenation pattern has been}%
\typeout{** loaded for the language `#1'. Using the pattern for}%
\typeout{** the default language instead.}%
\else
\language=\csname l@#1\endcsname
\fi
#2}}
\providecommand{\BIBdecl}{\relax}
\BIBdecl

\bibitem{tkgsurvey}
\BIBentryALTinterwordspacing
B.~Cai, Y.~Xiang, L.~Gao, H.~Zhang, Y.~Li, and J.~Li, ``Temporal knowledge graph completion: A survey,'' in \emph{International Joint Conference on Artificial Intelligence}, 2022. [Online]. Available: \url{https://api.semanticscholar.org/CorpusID:246063616}
\BIBentrySTDinterwordspacing

\bibitem{zhouwt}
\BIBentryALTinterwordspacing
W.-T. Zhou, Z.~Kang, L.~Tian, and Y.~Su, ``Intensity-free convolutional temporal point process: Incorporating local and global event contexts,'' \emph{Information Sciences}, vol. 646, p. 119318, 2023. [Online]. Available: \url{https://www.sciencedirect.com/science/article/pii/S0020025523009039}
\BIBentrySTDinterwordspacing

\bibitem{llmtkgr}
\BIBentryALTinterwordspacing
S.~Xiong, A.~Payani, R.~Kompella, and F.~Fekri, ``Large language models can learn temporal reasoning,'' in \emph{The 62nd Annual Meeting of the Association for Computational Linguistics, {ACL} 2024, Bangkok, Thailand, August 11–16, 2024}, 2024. [Online]. Available: \url{https://doi.org/10.48550/arXiv.2401.06853}
\BIBentrySTDinterwordspacing

\bibitem{snet}
Z.~Cai, Z.~He, X.~Guan, and Y.~Li, ``Collective data-sanitization for preventing sensitive information inference attacks in social networks,'' \emph{IEEE Transactions on Dependable and Secure Computing}, vol.~15, no.~4, pp. 577--590, 2018.

\bibitem{Iot}
Z.~Cai and X.~Zheng, ``A private and efficient mechanism for data uploading in smart cyber-physical systems,'' \emph{IEEE Transactions on Network Science and Engineering}, vol.~7, no.~2, pp. 766--775, 2020.

\bibitem{regcn}
Z.~Li, X.~Jin, W.~Li, S.~Guan, J.~Guo, H.~Shen, Y.~Wang, and X.~Cheng, ``Temporal knowledge graph reasoning based on evolutional representation learning,'' in \emph{Proceedings of the 44th International ACM SIGIR Conference on Research and Development in Information Retrieval}, 2021, pp. 408--417.

\bibitem{tirgn}
Y.~Li, S.~Sun, and J.~Zhao, ``Tirgn: Time-guided recurrent graph network with local-global historical patterns for temporal knowledge graph reasoning,'' in \emph{Proceedings of the Thirty-First International Joint Conference on Artificial Intelligence, {IJCAI} 2022, Vienna, Austria, 23-29 July 2022}, 2022, pp. 2152--2158.

\bibitem{cygnet}
C.~Zhu, M.~Chen, C.~Fan, G.~Cheng, and Y.~Zhang, ``Learning from history: Modeling temporal knowledge graphs with sequential copy-generation networks,'' in \emph{Proceedings of the AAAI Conference on Artificial Intelligence}, vol.~35, no.~5, 2021, pp. 4732--4740.

\bibitem{cenet}
Y.~Xu, J.~Ou, H.~Xu, and L.~Fu, ``Temporal knowledge graph reasoning with historical contrastive learning,'' in \emph{Proceedings of the AAAI Conference on Artificial Intelligence}, 2023.

\bibitem{renet}
W.~Jin, M.~Qu, X.~Jin, and X.~Ren, ``Recurrent event network: Autoregressive structure inferenceover temporal knowledge graphs,'' in \emph{Proceedings of the 2020 Conference on Empirical Methods in Natural Language Processing (EMNLP)}.\hskip 1em plus 0.5em minus 0.4em\relax Online: Association for Computational Linguistics, Nov. 2020, pp. 6669--6683.

\bibitem{cen}
Z.~Li, S.~Guan, X.~Jin, W.~Peng, Y.~Lyu, Y.~Zhu, L.~Bai, W.~Li, J.~Guo, and X.~Cheng, ``Complex evolutional pattern learning for temporal knowledge graph reasoning,'' in \emph{Proceedings of the 60th Annual Meeting of the Association for Computational Linguistics (Volume 2: Short Papers)}.\hskip 1em plus 0.5em minus 0.4em\relax Dublin, Ireland: Association for Computational Linguistics, May 2022, pp. 290--296.

\bibitem{RETIA}
\BIBentryALTinterwordspacing
K.~Liu, F.~Zhao, G.~Xu, X.~Wang, and H.~Jin, ``{RETIA:} relation-entity twin-interact aggregation for temporal knowledge graph extrapolation,'' in \emph{39th {IEEE} International Conference on Data Engineering, {ICDE} 2023, Anaheim, CA, USA, April 3-7, 2023}.\hskip 1em plus 0.5em minus 0.4em\relax {IEEE}, 2023, pp. 1761--1774. [Online]. Available: \url{https://doi.org/10.1109/ICDE55515.2023.00138}
\BIBentrySTDinterwordspacing

\bibitem{rpc}
\BIBentryALTinterwordspacing
K.~Liang, L.~Meng, M.~Liu, Y.~Liu, W.~Tu, S.~Wang, S.~Zhou, and X.~Liu, ``Learn from relational correlations and periodic events for temporal knowledge graph reasoning,'' in \emph{Proceedings of the 46th International ACM SIGIR Conference on Research and Development in Information Retrieval}, ser. SIGIR '23.\hskip 1em plus 0.5em minus 0.4em\relax New York, NY, USA: Association for Computing Machinery, 2023, p. 1559–1568. [Online]. Available: \url{https://doi.org/10.1145/3539618.3591711}
\BIBentrySTDinterwordspacing

\bibitem{l2tkg}
\BIBentryALTinterwordspacing
M.~Zhang, Y.~Xia, Q.~Liu, S.~Wu, and L.~Wang, ``Learning latent relations for temporal knowledge graph reasoning,'' in \emph{Proceedings of the 61st Annual Meeting of the Association for Computational Linguistics (Volume 1: Long Papers), {ACL} 2023, Toronto, Canada, July 9-14, 2023}, A.~Rogers, J.~L. Boyd{-}Graber, and N.~Okazaki, Eds.\hskip 1em plus 0.5em minus 0.4em\relax Association for Computational Linguistics, 2023, pp. 12\,617--12\,631. [Online]. Available: \url{https://doi.org/10.18653/v1/2023.acl-long.705}
\BIBentrySTDinterwordspacing

\bibitem{hgls}
\BIBentryALTinterwordspacing
------, ``Learning long- and short-term representations for temporal knowledge graph reasoning,'' in \emph{Proceedings of the ACM Web Conference 2023}, ser. WWW '23.\hskip 1em plus 0.5em minus 0.4em\relax New York, NY, USA: Association for Computing Machinery, 2023, p. 2412–2422. [Online]. Available: \url{https://doi.org/10.1145/3543507.3583242}
\BIBentrySTDinterwordspacing

\bibitem{logCL}
W.~Chen, H.~Wan, Y.~Wu, S.~Zhao, J.~Chen, Y.~Li, and Y.~Lin, ``Local-global history-aware contrastive learning for temporal knowledge graph reasoning,'' in \emph{40th {IEEE} International Conference on Data Engineering, {ICDE} 2024, Utrecht, Netherlands, May 13-16, 2024}.\hskip 1em plus 0.5em minus 0.4em\relax {IEEE}, 2024.

\bibitem{icews}
\BIBentryALTinterwordspacing
E.~Boschee, J.~Lautenschlager, S.~O'Brien, S.~Shellman, J.~Starz, and M.~Ward, ``{ICEWS Coded Event Data},'' 2015. [Online]. Available: \url{https://doi.org/10.7910/DVN/28075}
\BIBentrySTDinterwordspacing

\bibitem{tian}
L.~Tian, J.~Zhang, J.~Zhang, W.~Zhou, and X.~Zhou, ``Knowledge graph survey: representation, construction, reasoning and knowledge hypergraph theory,'' \emph{Journal of Computer Applications}, vol.~41, no.~8, p. 2161, 2021.

\bibitem{fb15k}
K.~Toutanova, D.~Chen, P.~Pantel, H.~Poon, P.~Choudhury, and M.~Gamon, ``Representing text for joint embedding of text and knowledge bases,'' in \emph{Proceedings of the 2015 conference on empirical methods in natural language processing}, 2015, pp. 1499--1509.

\bibitem{conve}
\BIBentryALTinterwordspacing
T.~Dettmers, M.~Pasquale, S.~Pontus, and S.~Riedel, ``Convolutional 2d knowledge graph embeddings,'' in \emph{Proceedings of the 32th AAAI Conference on Artificial Intelligence}, February 2018, pp. 1811--1818. [Online]. Available: \url{https://arxiv.org/abs/1707.01476}
\BIBentrySTDinterwordspacing

\bibitem{transe}
A.~Bordes, N.~Usunier, A.~Garcia-Duran, J.~Weston, and O.~Yakhnenko, ``Translating embeddings for modeling multi-relational data,'' in \emph{Advances in Neural Information Processing Systems}, C.~Burges, L.~Bottou, M.~Welling, Z.~Ghahramani, and K.~Weinberger, Eds., vol.~26.\hskip 1em plus 0.5em minus 0.4em\relax Curran Associates, Inc., 2013.

\bibitem{distmult}
\BIBentryALTinterwordspacing
B.~Yang, W.~Yih, X.~He, J.~Gao, and L.~Deng, ``Embedding entities and relations for learning and inference in knowledge bases,'' in \emph{3rd International Conference on Learning Representations, {ICLR} 2015, San Diego, CA, USA, May 7-9, 2015, Conference Track Proceedings}, Y.~Bengio and Y.~LeCun, Eds., 2015. [Online]. Available: \url{http://arxiv.org/abs/1412.6575}
\BIBentrySTDinterwordspacing

\bibitem{complex}
T.~Trouillon, J.~Welbl, S.~Riedel, E.~Gaussier, and G.~Bouchard, ``Complex embeddings for simple link prediction,'' in \emph{Proceedings of The 33rd International Conference on Machine Learning}, ser. Proceedings of Machine Learning Research, M.~F. Balcan and K.~Q. Weinberger, Eds., vol.~48.\hskip 1em plus 0.5em minus 0.4em\relax New York, New York, USA: PMLR, 20--22 Jun 2016, pp. 2071--2080.

\bibitem{rotate}
\BIBentryALTinterwordspacing
Z.~Sun, Z.~Deng, J.~Nie, and J.~Tang, ``Rotate: Knowledge graph embedding by relational rotation in complex space,'' in \emph{7th International Conference on Learning Representations, {ICLR} 2019, New Orleans, LA, USA, May 6-9, 2019}.\hskip 1em plus 0.5em minus 0.4em\relax OpenReview.net, 2019. [Online]. Available: \url{https://openreview.net/forum?id=HkgEQnRqYQ}
\BIBentrySTDinterwordspacing

\bibitem{convtranse}
C.~Shang, Y.~Tang, J.~Huang, J.~Bi, X.~He, and B.~Zhou, ``End-to-end structure-aware convolutional networks for knowledge base completion,'' in \emph{Proceedings of the AAAI Conference on Artificial Intelligence}, vol.~33, no.~01, 2019, pp. 3060--3067.

\bibitem{gcn}
\BIBentryALTinterwordspacing
T.~N. Kipf and M.~Welling, ``Semi-supervised classification with graph convolutional networks,'' in \emph{5th International Conference on Learning Representations, {ICLR} 2017, Toulon, France, April 24-26, 2017, Conference Track Proceedings}.\hskip 1em plus 0.5em minus 0.4em\relax OpenReview.net, 2017. [Online]. Available: \url{https://openreview.net/forum?id=SJU4ayYgl}
\BIBentrySTDinterwordspacing

\bibitem{rgcn}
M.~Schlichtkrull, T.~N. Kipf, P.~Bloem, R.~v.~d. Berg, I.~Titov, and M.~Welling, ``Modeling relational data with graph convolutional networks,'' in \emph{European semantic web conference}.\hskip 1em plus 0.5em minus 0.4em\relax Springer, 2018, pp. 593--607.

\bibitem{compgcn}
S.~Vashishth, S.~Sanyal, V.~Nitin, and P.~P. Talukdar, ``Composition-based multi-relational graph convolutional networks,'' in \emph{8th International Conference on Learning Representations, {ICLR} 2020, Addis Ababa, Ethiopia, April 26-30, 2020}.\hskip 1em plus 0.5em minus 0.4em\relax OpenReview.net, 2020.

\bibitem{gat}
\BIBentryALTinterwordspacing
P.~Veli{\v{c}}kovi{\'{c}}, G.~Cucurull, A.~Casanova, A.~Romero, P.~Li{\`{o}}, and Y.~Bengio, ``{Graph Attention Networks},'' \emph{International Conference on Learning Representations}, 2018, accepted as poster. [Online]. Available: \url{https://openreview.net/forum?id=rJXMpikCZ}
\BIBentrySTDinterwordspacing

\bibitem{KBGAT}
\BIBentryALTinterwordspacing
D.~Nathani, J.~Chauhan, C.~Sharma, and M.~Kaul, ``Learning attention-based embeddings for relation prediction in knowledge graphs,'' in \emph{Proceedings of the 57th Annual Meeting of the Association for Computational Linguistics}, A.~Korhonen, D.~Traum, and L.~M{\`a}rquez, Eds.\hskip 1em plus 0.5em minus 0.4em\relax Florence, Italy: Association for Computational Linguistics, Jul. 2019, pp. 4710--4723. [Online]. Available: \url{https://aclanthology.org/P19-1466}
\BIBentrySTDinterwordspacing

\bibitem{tilp}
\BIBentryALTinterwordspacing
S.~Xiong, Y.~Yang, F.~Fekri, and J.~C. Kerce, ``{TILP:} differentiable learning of temporal logical rules on knowledge graphs,'' in \emph{The Eleventh International Conference on Learning Representations, {ICLR} 2023, Kigali, Rwanda, May 1-5, 2023}.\hskip 1em plus 0.5em minus 0.4em\relax OpenReview.net, 2023. [Online]. Available: \url{https://openreview.net/pdf?id=\_X12NmQKvX}
\BIBentrySTDinterwordspacing

\bibitem{teilp}
\BIBentryALTinterwordspacing
S.~Xiong, Y.~Yang, A.~Payani, J.~C. Kerce, and F.~Fekri, ``{TEILP:} time prediction over knowledge graphs via logical reasoning,'' in \emph{Thirty-Eighth {AAAI} Conference on Artificial Intelligence, {AAAI} 2024, Thirty-Sixth Conference on Innovative Applications of Artificial Intelligence, {IAAI} 2024, Fourteenth Symposium on Educational Advances in Artificial Intelligence, {EAAI} 2014, February 20-27, 2024, Vancouver, Canada}, M.~J. Wooldridge, J.~G. Dy, and S.~Natarajan, Eds.\hskip 1em plus 0.5em minus 0.4em\relax {AAAI} Press, 2024, pp. 16\,112--16\,119. [Online]. Available: \url{https://doi.org/10.1609/aaai.v38i14.29544}
\BIBentrySTDinterwordspacing

\bibitem{tilr}
Y.~Yang, S.~Xiong, J.~C. Kerce, and F.~Fekri, ``Temporal inductive logic reasoning over hypergraphs,'' in \emph{The 33rd International Joint Conference on Artificial Intelligence, {IJCAI} 2024, Jeju, Korea, August 3-9, 2024}, 2024.

\bibitem{xerte}
Z.~Han, P.~Chen, Y.~Ma, and V.~Tresp, ``Explainable subgraph reasoning for forecasting on temporal knowledge graphs,'' in \emph{International Conference on Learning Representations}, 2021.

\bibitem{ghnn}
Z.~Han, Y.~Ma, Y.~Wang, S.~G{\"{u}}nnemann, and V.~Tresp, ``Graph hawkes neural network for forecasting on temporal knowledge graphs,'' in \emph{Conference on Automated Knowledge Base Construction, {AKBC} 2020, Virtual, June 22-24, 2020}, D.~Das, H.~Hajishirzi, A.~McCallum, and S.~Singh, Eds., 2020.

\bibitem{tango}
Z.~Han, Z.~Ding, Y.~Ma, Y.~Gu, and V.~Tresp, ``Learning neural ordinary equations for forecasting future links on temporal knowledge graphs,'' in \emph{Proceedings of the 2021 Conference on Empirical Methods in Natural Language Processing}, 2021, pp. 8352--8364.

\bibitem{titer}
H.~Sun, J.~Zhong, Y.~Ma, Z.~Han, and K.~He, ``Timetraveler: Reinforcement learning for temporal knowledge graph forecasting,'' in \emph{Proceedings of the 2021 Conference on Empirical Methods in Natural Language Processing}, 2021, pp. 8306--8319.

\bibitem{cluster}
Z.~Li, X.~Jin, S.~Guan, W.~Li, J.~Guo, Y.~Wang, and X.~Cheng, ``Search from history and reason for future: Two-stage reasoning on temporal knowledge graphs,'' in \emph{Proceedings of the 59th Annual Meeting of the Association for Computational Linguistics and the 11th International Joint Conference on Natural Language Processing (Volume 1: Long Papers)}, 2021, pp. 4732--4743.

\bibitem{TLogic}
Y.~Liu, Y.~Ma, M.~Hildebrandt, M.~Joblin, and V.~Tresp, ``Tlogic: Temporal logical rules for explainable link forecasting on temporal knowledge graphs,'' in \emph{Thirty-Sixth {AAAI} Conference on Artificial Intelligence, {AAAI} 2022, Thirty-Fourth Conference on Innovative Applications of Artificial Intelligence, {IAAI} 2022, The Twelveth Symposium on Educational Advances in Artificial Intelligence, {EAAI} 2022 Virtual Event, February 22 - March 1, 2022}.\hskip 1em plus 0.5em minus 0.4em\relax {AAAI} Press, 2022, pp. 4120--4127.

\bibitem{pleasing}
\BIBentryALTinterwordspacing
J.~Zhang, M.~Sun, Q.~Huang, and L.~Tian, ``Pleasing: Exploring the historical and potential events for temporal knowledge graph reasoning,'' \emph{Neural Networks}, p. 106516, 2024. [Online]. Available: \url{https://www.sciencedirect.com/science/article/pii/S0893608024004404}
\BIBentrySTDinterwordspacing

\bibitem{lms}
\BIBentryALTinterwordspacing
J.~Zhang, B.~Hui, C.~Mu, and L.~Tian, ``Learning multi-graph structure for temporal knowledge graph reasoning,'' \emph{Expert Systems with Applications}, vol. 255, p. 124561, 2024. [Online]. Available: \url{https://www.sciencedirect.com/science/article/pii/S0957417424014283}
\BIBentrySTDinterwordspacing

\bibitem{hismatch}
\BIBentryALTinterwordspacing
Z.~Li, Z.~Hou, S.~Guan, X.~Jin, W.~Peng, L.~Bai, Y.~Lyu, W.~Li, J.~Guo, and X.~Cheng, ``Hismatch: Historical structure matching based temporal knowledge graph reasoning,'' in \emph{Findings of the Association for Computational Linguistics: {EMNLP} 2022, Abu Dhabi, United Arab Emirates, December 7-11, 2022}, Y.~Goldberg, Z.~Kozareva, and Y.~Zhang, Eds.\hskip 1em plus 0.5em minus 0.4em\relax Association for Computational Linguistics, 2022, pp. 7328--7338. [Online]. Available: \url{https://doi.org/10.18653/v1/2022.findings-emnlp.542}
\BIBentrySTDinterwordspacing

\bibitem{gdelt}
K.~Leetaru and P.~A. Schrodt, ``Gdelt: Global data on events, location, and tone,'' \emph{ISA Annual Convention}, 2013.

\bibitem{evokg}
N.~Park, F.~Liu, P.~Mehta, D.~Cristofor, C.~Faloutsos, and Y.~Dong, ``Evokg: Jointly modeling event time and network structure for reasoning over temporal knowledge graphs,'' in \emph{Proceedings of the Fifteenth ACM International Conference on Web Search and Data Mining}, 2022, pp. 794--803.

\end{thebibliography}
\end{document}